\documentclass{article}
\usepackage{ijcai22}

\usepackage{times}
\usepackage{soul}
\usepackage{url}
\usepackage[hidelinks]{hyperref}
\usepackage[utf8]{inputenc}
\usepackage[small]{caption}
\usepackage{graphicx}
\usepackage{amsmath}
\usepackage{amsfonts}
\usepackage{amsthm}
\usepackage{booktabs}
\usepackage{algorithm}
\usepackage{amsfonts}
\usepackage{algorithmic}
\usepackage{booktabs}
\usepackage{multirow}
\usepackage{arydshln}
\usepackage{subfigure}
\usepackage{color}

\title{Data-Free Adversarial Knowledge Distillation for Graph Neural Networks}

\author{
Yuanxin Zhuang$^1$
\and
Lingjuan Lyu$^2$\and
Chuan Shi$^{1}$\footnote{Corresponding Author.}\and
Carl Yang$^3$\And
Lichao Sun$^4$
\affiliations
$^1$Beijing University of Posts and Telecommunications\\
$^2$Sony AI\quad
$^3$Emory University\quad
$^4$Lehigh University
\emails
\{zhuangyuanxin, shichuan\}@bupt.edu.cn,\\
lingjuan.lv@sony.com,
j.carlyang@emory.edu,
lis221@lehigh.edu
}

\begin{document}

\maketitle

\begin{abstract}
Graph neural networks (GNNs) have been widely used in modeling graph structured data, owing to its impressive performance in a wide range of practical applications. Recently, knowledge distillation (KD) for GNNs has enabled remarkable progress in graph model compression and knowledge transfer. However, most of the existing KD methods require a large volume of real data, which are not readily available in practice, and may preclude their applicability in scenarios where the teacher model is trained on rare or hard to acquire datasets. To address this problem, we propose the first end-to-end framework for data-free adversarial knowledge distillation on graph structured data (DFAD-GNN). To be specific, our DFAD-GNN employs a generative adversarial network, which mainly consists of three components: a pre-trained teacher model and a student model are regarded as two discriminators, and a generator is utilized for deriving training graphs to distill knowledge from the teacher model into the student model. Extensive experiments on various benchmark models and six representative datasets demonstrate that our DFAD-GNN significantly surpasses state-of-the-art data-free baselines in the graph classification task. 
% Code is available at 
% \small{https://anonymous.4open.science/r/DF-GNNs-EC75}.
\end{abstract}

% \section{Introduction}

\section{Introduction}

\noindent An increasing number of machine learning tasks require dealing with a large amount of graph data, which capture rich and complex relationships among potentially billions of elements. Graph Neural Networks (GNNs) have become an effective way to address the graph learning problem by converting the graph data into a low dimensional space while keeping both the structural and property information to the maximum extent. Recently, the rapid evolution of GNNs has led to a growing number of new architectures as well as novel applications \cite{lu2021learning,sun2018adversarial,wu2021fedgnn}.

However, training a powerful GNN often requires heavy computation and storage. Hence it is hard to deploy them into resource-constrained devices, such as mobile phones. There has been a large literature \cite{bahri2021binary} aiming to compress and speed-up the cumbersome GNNs into lightweight ones. Among those methods, knowledge distillation \cite{hinton2015distilling} is one of the most popular paradigms for learning a portable student model from the pre-trained complicated teacher by directly imitating its outputs.

Knowledge distillation (KD) is proposed by Hinton et al. \cite{hinton2015distilling} for supervising the training of a compact yet efficient student model by capturing and transferring the knowledge from a large complicated teacher model. KD has received significant attention from the community in the recent years \cite{yang2021extract,gou2021knowledge}. Despite its successes, KD in its classical form has a critical limitation. It assumes that the real training data is still available in the distillation phase. However, in practice, the original training data is often unavailable due to privacy concerns. Besides, many large models are trained on millions or even billions of graphs \cite{lu2021learning}. While the pre-trained models might be made available for the community at large, making training data available also poses a lot of technical and policy challenges. 

An effective way to avert the above-mentioned issue is using the synthetic graphs, 
i.e., data-free knowledge distillation \cite{lopes2017data,liu2021data}. Just as ``data free'' implies, there is no training data. Instead, the data is reversely generated from the pre-trained models. Data-free distillation has received a lot of attention in the field of computer vision \cite{fang2019data,lopes2017data,fang2021up}, which is however rarely been explored in graph mining. Note that Deng et al. \cite{Deng2021GraphFreeKD} have made some pilot studies on this problem and proposed graph-free knowledge distillation (GFKD). Unfortunately, GFKD is not an end-to-end approach. It only takes the fixed teacher model into account, ignoring the information from the student model when generating graphs. Moreover, their %generation constraints are empirically designed
method is based on the assumption that an appropriate graph usually has a high degree of confidence in the teacher model. In fact, the model maps the graphs from the data space to a very small output space, thus losing a large amount of information. Therefore, these generated graphs are not very useful for distilling the teacher model efficiently which leads to an unsatisfactory performance.

In this work, we propose a novel data-free adversarial knowledge distillation framework for GNNs (DFAD-GNN). DFAD-GNN uses a knowledge distillation method based
on GAN \cite{goodfellow2014generative}. As illustrated in Figure \ref{fig:model}, DFAD-GNN contains one generator and two discriminators: one fixed discriminator is the pre-trained teacher model, the other is the compact student model that we aim to learn. The generator generates graphs to help transfer teachers’ knowledge to students. Unlike previous work \cite{Deng2021GraphFreeKD}, our generator can fully utilize both the intrinsic statistics from the pre-trained teacher model and the customizable information from the student model, which can help generate high quality and diverse training data to improve the student model's generalizability.
The contributions of our proposed framework can be summarized as follows:
1) We study a valuable yet intractable problem: how to distill a portable and efficient student model from a pre-trained teacher model when the original training data is not available; 2) We propose a novel data-free adversarial knowledge distillation framework for GNNs (DFAD-GNN) in order to train a compact student model using the generated graphs and a fixed teacher model. To the best of our knowledge, DFAD-GNN is the first end-to-end framework for data-free knowledge distillation on graph structured data; and 3) Extensive experiments demonstrate that our proposed framework significantly outperforms existing state-of-the-art data-free methods. DFAD-GNN can successfully distill a student model with 81.8\%-94.7\% accuracy of the teacher model on all six datasets. %across different domains. 

\section{Preliminary and Related Work}

\subsection{Graph Neural Networks}

Graph Neural Networks (GNNs) have received considerable attention for a wide variety of tasks \cite{wu2020comprehensive,zhou2020graph}. Generally, GNN models can be unified by a neighborhood aggregation or message passing schema \cite{gilmer2017neural}, where the representation of each node is learned by iteratively aggregating the embeddings (“message”) of its neighbors.

As one of the most influential GNN models, Graph Convolutional Network (GCN) \cite{kipf2016semi} performs a linear approximation to graph convolutions. Graph Attention Network (GAT) \cite{velivckovic2017graph} introduces an attention mechanism that allows weighing nodes in the neighborhood differently during the aggregation step. GraphSAGE \cite{hamilton2017inductive} is a comprehensive improvement on the original GCN which replaced full graph Laplacian with learnable aggregation functions. Graph Isomorphism Network (GIN) \cite{xu2018powerful} uses a simple but expressive injective multiset function for neighbor aggregation. These GNNs above will be employed as our teacher models and student models in the experiments.

% DELETE
% Formally, the k-th layer of a GNN is:
% \begin{equation}
% \begin{aligned}
% &h_{v}^{(k)}=\operatorname{COMBINE}^{(k)}\left(h_{v}^{(k-1)},\right.\\
% &\left.\text{AGGREGATE }^{(k)}\left(\left\{\left(h_{v}^{(k-1)}, h_{u}^{(k-1)}\right): u \in \mathcal{N}(v)\right\}\right)\right),
% \end{aligned}
% \end{equation}
% where $h_{v}^{(k)}$ is the representation of node $v$ at the $k$-th iteration/layer, and $\mathcal{N}(v)$ is a set neighbors of $v$. We initialize $h_{v}^{(0)}=F_{v}$ which is the node feature.

% To obtain the entire graph’s representation $h_{G}$, the READOUT function pools node features from the final iteration $K$,
% \begin{equation}
% h_{G}=\operatorname{READOUT}\left(\left\{h_{v}^{(K)} \mid v \in G\right\}\right).
% \end{equation}
% READOUT is a permutation-invariant function, such as averaging or a more sophisticated graph-level pooling function.

\subsection{Knowledge Distillation}

Knowledge distillation (KD) aims to 
transfer the knowledge of a (larger) teacher model to a (smaller) student model \cite{hinton2015distilling,wu2022communication}. It was originally introduced to reduce the size of models deployed on devices with limited computational resources. Since then, this line of research has attracted a lot of attention \cite{gou2021knowledge}. Recently, there are a few attempts that try to combine knowledge distillation with graph convolutional networks (GCNs). Yang et al. \cite{yang2021extract} proposed a knowledge distillation framework which can extract the knowledge of an arbitrary teacher model and inject it into a well-designed student model. Jing et al.~\cite{jing2021amalgamating} proposed to learn a lightweight student GNN that masters the complete set of expertise of multiple heterogeneous teachers. These works aim to improve the performance of student models on semi-supervised node classification task, rather than the graph classification task we considered in this work. Furthermore, although the above-mentioned methods obtained promising results, they cannot be effectively launched without the original training dataset. In practice, the training dataset could be unavailable for some reasons, e.g. transmission limitations, privacy, etc. 
Therefore, it is necessary to consider a data-free approach for compressing neural networks.

Techniques addressing data-free knowledge distillation have relied on training a generative model to synthesize fake data. One recent work named graph-free KD (GFKD) \cite{Deng2021GraphFreeKD} has proposed a data-free knowledge distillation for graph neural network. GFKD learns graph topology structures for knowledge distillation by modeling them with a multinomial distribution. The training processes include two independent steps: (1) first, it uses a pre-trained teacher model to generate fake graphs; (2) afterwards, it uses these fake graphs to distill knowledge into the compact student model. However, those fake graphs are optimized by an invariant teacher model without considering the student model. Therefore, they are not very useful for distilling the student model efficiently. In order to generate high quality diverse training data to improve the student model's generalizability, we propose a data-free adversarial knowledge distillation framework for GNNs (DFAD-GNN). Our generator is trained end-to-end which not only uses the pre-trained teacher’s intrinsic statistics but also obtains the discrepancy between teacher model and student model. We remark that the key differences between GFKD and our model lies in the training process and the generator. 

\subsection{Graph Generation}

Data-free knowledge distillation involves the generation of training data. Motivated by the power of Generative Adversarial Networks (GANs) \cite{goodfellow2014generative}, researchers have used them for generating graphs. Bojchevski et al. proposed NetGAN \cite{bojchevski2018netgan}, which uses the GAN framework to generate random walks on graphs. De Cao and Kipf proposed MolGAN \cite{de2018molgan}, which generates molecular graphs using a simple multi-layer perceptron (MLP).

In this work, we build on a min-max game between two adversaries who try to optimize opposite loss functions. This approach is analogous to the optimization performed in GANs to train the generator and discriminator. The key difference is that GANs are generally trained to recover an underlying fixed data distribution. However, our generator chases a moving target: the distribution of data which is most indicative of the discrepancies of the current student model and its teacher model.

\begin{figure}[t]
\centering
\includegraphics[width=0.9\columnwidth]{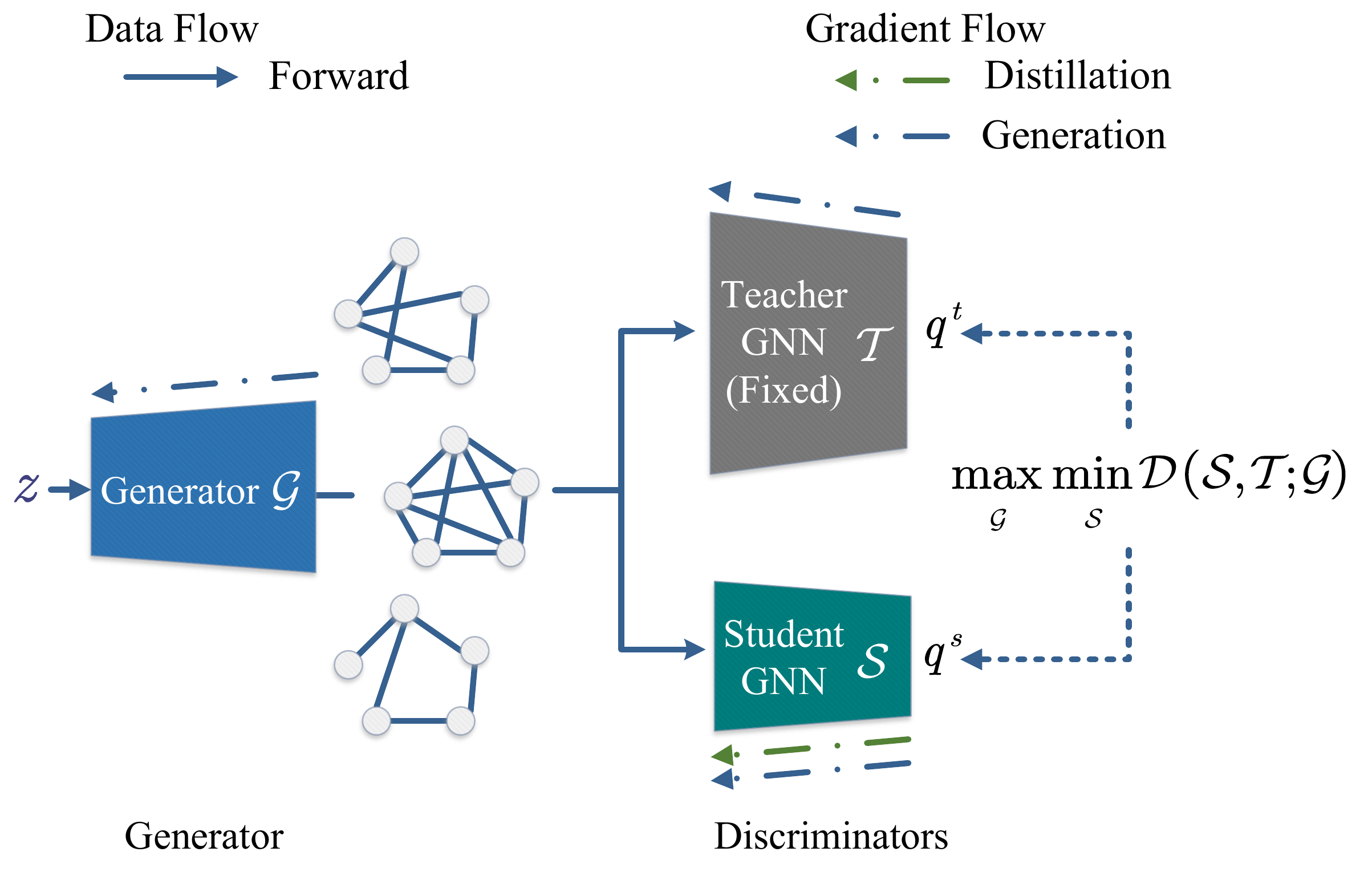} 
\caption{The framework of DFAD-GNN.} 
\label{fig:model}
\end{figure}

\section{DFAD-GNN Framework}
As shown in Figure~\ref{fig:model}, DFAD-GNN mainly consists of three components: one generator and two discriminators. One fixed discriminator is the pre-trained teacher model $\mathcal{T}$, the other is the compact student model $\mathcal{S}$ that we aim to learn. More specifically, the generator $\mathcal{G}$ takes samples $z$ from a prior distribution and generates fake graphs. Then the generated graphs are used to train a student model under the supervision of the teacher model.

\subsection{Generator}
The generator $\mathcal{G}$ is used to synthesize fake graphs that maximize the disagreement between the teacher $\mathcal{T}$ and the student $\mathcal{S}$. $\mathcal{G}$ takes $D$-dimensional vectors ${z} \in \mathbb{R}^{D}$ sampled from a standard normal distribution ${z} \sim \mathcal{N}(0, I)$ and outputs graphs. For each $z$, $\mathcal{G}$ outputs an object: ${F} \in \mathbb{R}^{N \times T}$ that defines node features, where $N$ is the node number and $T$ is node feature dimension. Then we calculate adjacency matrix ${{A}}$ as follows:

\begin{equation}
{{A}}=\sigma\left({F} {F}^{\top}\right),
\end{equation}
where $\sigma(\cdot)$ is the logistic sigmoid function. We transform the range of $A$ from $[0,1]$ to $\{0,1\}$ with a threshold $\tau$. If the element in $A$ is larger than $\tau$, it is set as 1, otherwise 0.

The loss function used for $\mathcal{G}$ is the same as that used for $\mathcal{S}$, except that the goal is to maximize it. 
% \st{We formulate this problem as an adversarial game in which $\mathcal{G}$ and $\mathcal{S}$ compete to respectively maximize and minimize the same function. }
In other words, the student is trained to match the teacher’s predictions and the generator is trained to generate difficult graphs for the student. The training process can be formulated as:

\begin{equation}
\max _{\mathcal{G}} \min _{\mathcal{S}} \mathbb{E}_{z \sim \mathcal{N}(0,1)}[\mathcal{D}(\mathcal{T}(\mathcal{G}(z)), \mathcal{S}(\mathcal{G}(z)))],
\end{equation}
where $\mathcal{D}\left(\cdot \right)$ indicates the %differences
discrepancy between the teacher $\mathcal{T}$ and the student $\mathcal{S}$. If generator keeps generating simple and duplicate graphs, student model will fit these graphs, resulting in a very low model discrepancy between student model and teacher model. In this case, the generator is forced to generate difficult and different graphs to enlarge the discrepancy.

\subsection{Adversarial Distillation}

% \st{As aforementioned, the generator $\mathcal{G}\left(z , \theta^{g}\right)$ is used to generate graphs. The student model $\mathcal{S}\left(x , \theta^{s}\right)$, together with the teacher model $\mathcal{T}\left(x , \theta^{t}\right)$ are jointly viewed as the discriminator to measure the discrepancy $\mathcal{D}\left(\mathcal{T} , \mathcal{S};\mathcal{G}\right)$. }

Overall, the adversarial training process consists of two stages: the distillation stage that minimizes the discrepancy $\mathcal{D}$; and the generation stage that maximizes the discrepancy $\mathcal{D}$, as shown in Figure \ref{fig:model}. We detail each stage as follows.

\begin{algorithm}[t]
\caption{DFAD-GNN}
\label{alg:algorithm}
\textbf{Input}: A pre-trained teacher model,$\mathcal{T}\left(X ; \theta^{t}\right)$\\
% \textbf{Parameter}: Optional list of parameters\\
\textbf{Output}: A comparable student model,$\mathcal{S}\left(X ; \theta^{s}\right)$\\
\begin{algorithmic}[1] %[1] enables line numbers
\STATE Randomly initialize a student model $\mathcal{S}\left(X ; \theta^{s}\right)$ and a generator $\mathcal{G}\left(z ; \theta^{g}\right)$
\FOR {Epochs}
\STATE // Distillation Stage
\FOR{$k$ steps}
\STATE Generate graphs $X$ from $z$ with $\mathcal{G}\left(z ; \theta^{g}\right)$
\STATE Calculate model discrepancy with $\mathcal{L}_{D I S}$
\STATE Update $\theta^{s}$ to minimize discrepancy with $\nabla_{\theta^{s}} \mathcal{D}$
% $\nabla_{\theta^{s}} \mathcal{D}\left(\mathcal{T}, \mathcal{S} ; \mathcal{G}\right)$
\ENDFOR \\
\STATE //Generation Stage
\STATE Generate graphs $X$ from $z$ with $\mathcal{G}\left(z ; \theta^{g}\right)$;
\STATE Calculate negative discrepancy with $\mathcal{L}_{G E N}$
\STATE Update $\theta^{g}$ to maximize discrepancy with $\nabla_{\theta^{g}}-\mathcal{D}$
\ENDFOR
% \STATE \textbf{return} solution
\end{algorithmic}
\end{algorithm}

\subsubsection{Distillation Stage}

In this stage, we fix the generator $\mathcal{G}$ and only update the student $\mathcal{S}$ in the discriminator. We sample a batch of random noises ${z}$ and construct fake graphs with generator $\mathcal{G}$. Then each graph ${X}$ is fed to both the teacher and the student model to produce the output $q^{t}$ and $q^{s}$, where $q$ is a vector indicating the scores of different categories.

% The choice of loss is of paramount importance to a successful distillation. 
In our approach, the choice of loss involves similar factors to those outlined in %research on 
GANs: multiple works have discussed the problem of vanishing gradients as the discriminator becomes strong in case of GAN training \cite{arjovsky2017towards}. Most prior work in model distillation optimized over the KullbackLeibler Divergence (KLD) and Mean Square Error (MSE) between the student and the teacher model. However, as the student model matches more closely the
teacher model, these two loss functions tend to suffer from vanishing gradients. Specifically, back-propagating such vanishing gradients through the generator
can harm its learning. For our approach, we minimize the Mean Absolute Error (MAE) between $q^{t}$ and $q^{s}$, which provides stable gradients for the generator so that the vanishing gradients can be alleviated. In our experiment, we empirically find that this significantly improves student's performance over other possible losses. Now we can define the loss function for distillation stage as follows:
\begin{equation}
\mathcal{L}_{D I S} =\mathbb{E}_{z \sim p_{z}(z)}\left[\frac{1}{n}\left\|\mathcal{T}\left(\mathcal{G}(z)\right)-\mathcal{S}\left(\mathcal{G}(z)\right)\right\|_{1}\right].
\end{equation}

%Intuitively, this stage is very similar to KD, but the goals are slightly different. In KD, students can greedily learn from the soft targets produced by the teacher, as these targets are obtained from the real data \cite{hinton2015distilling} and contain useful knowledge for the specific task. However, in our setting, we have no access to any real data. The fake graphs synthesized by the generator are not guaranteed to be useful, especially at the beginning of training. As aforementioned, the generator is required to produce graphs to measure the model discrepancy between the teacher model and the student model. Another essential purpose of the distillation stage is to construct a better search space to force the generator to find new graphs.

\subsubsection{Generation Stage}

The goal of the generation stage is to push the generation of new graphs. In this stage, we fix the two discriminators and only update the generator. We encourage the generator to produce more confusing training graphs. The loss function used for generator is the same as for student except that the goal is to maximize it:

\begin{equation}
\begin{aligned}
\mathcal{L}_{G E N} = -\mathcal{L}_{D I S}.
\end{aligned}
\end{equation}

With the generation loss, the error first back-propagates through the discriminator (the teacher and the student model), then through the generator to optimize it.

\subsection{Optimization}

The whole distillation process is summarized in Algorithm \ref{alg:algorithm}. DFAD-GNN trains the student and the generator by iterating over the distillation stage and the generation stage. 
% \st{It begins with the distillation stage to minimize the discrepancy $\mathcal{D}\left(\mathcal{T}, \mathcal{S} ; \mathcal{G}\right)$. Then in the generation stage, we update the generator to maximize $\mathcal{D}\left(\mathcal{T}, \mathcal{S} ; \mathcal{G}\right)$. }
Based on the learning progress of the student model, the generator crafts new graphs to further estimate the model discrepancy. The competition in this adversarial game drives the generator to discover more knowledge. %It is essential to maintain stability in adversarial training. 
In the distillation stage, we update the student model for $k$ times so as to ensure its convergence. 
Note that compared with the conventional method GFKD~\cite{Deng2021GraphFreeKD}, the time complexity of DFAD-GNN mainly lies on the matrix multiplication of the generator, i.e., $O(TN^2)$ where $N$ is the number of nodes and $T$ is the node feature dimension. Although our method has higher time complexity, the performances are much better. In addition, because there are not too many nodes in most real world graph-level applications (usually less than 100), we remark that our complexity is acceptable in practice.

\section{Experiments}
\subsection{Datasets}
We adopt six graph classification benchmark datasets including three bioinformatics graph datasets, i.e., MUTAG, PTC\_MR, and PROTEINS, and three social network graph datasets, i.e., IMDB-BINARY, COLLAB, and REDDIT-BINARY. The statistics of these datasets are summarized in Table \ref{tab:data}. To remove the unwanted bias towards the training data, for all experiments on these datasets, we evaluate the model performance with a 10-fold cross validation setting, where the dataset split is based on the conventionally used training/test splits \cite{niepert2016learning,zhang2018end,xu2018powerful} with LIBSVM \cite{chang2011libsvm}. We report the average and standard deviation of validation accuracies across the 10 folds within the cross-validation.
\begin{table}[h]
\scalebox{0.90}{
\centering
\begin{tabular}{cccc}
\hline
Dataset         & \#Graphs & \#Classes & Avg\#Graph Size \\ \hline
MUTAG    & 188      & 2         & 17.93           \\
PTC\_MR  & 344      & 2         & 14.29           \\
PROTEINS & 1113     & 2         & 39.06           \\
IMDB-BINARY   & 1000     & 2         & 19.77           \\
COLLAB   & 5000     & 3         & 74.49           \\
REDDIT-BINARY & 2000     & 2         & 427.62         
\\ \hline
\end{tabular}}
\caption{Summary of datasets.}
\label{tab:data}
\end{table}

\begin{table*}[ht]
\centering
\scalebox{0.6}{
\begin{tabular}{ccc|cc|cc|cc|cc|cc}

\hline
Datasets &
  \multicolumn{2}{c|}{MUTAG} &
  \multicolumn{2}{c|}{PTC\_MR} &
  \multicolumn{2}{c|}{PROTEINS} &
  \multicolumn{2}{c|}{IMDB-BINARY} &
  \multicolumn{2}{c|}{COLLAB} &
  \multicolumn{2}{c}{REDDIT-Binary} \\ \hline
\multirow{2}{*}{Teacher} &
  \multicolumn{2}{c|}{GIN-5-128} &
  \multicolumn{2}{c|}{GIN-5-128} &
  \multicolumn{2}{c|}{GIN-5-128} &
  \multicolumn{2}{c|}{GIN-5-128} &
  \multicolumn{2}{c|}{GIN-5-128} &
  \multicolumn{2}{c}{GIN-5-128} \\
 &
  \multicolumn{2}{c|}{96.7±3.7} &
  \multicolumn{2}{c|}{75.0±3.5} &
  \multicolumn{2}{c|}{78.3±2.9} 
  &
  
  \multicolumn{2}{c|}{80.1±3.7} &
  \multicolumn{2}{c|}{83.5±1.2} &
  \multicolumn{2}{c}{92.2±1.2}\\ \hline
Student &
  GIN-5-32 &
  GIN-1-128 &
  GIN-5-32 &
  GIN-1-128 &
  GIN-5-32 &
  GIN-1-128&
  GIN-5-32 &
  GIN-1-128 &
  GIN-5-32 &
  GIN-1-128 &
  GIN-5-32 &
  GIN-1-128\\ 
 &
  (6.7\%$\times m$) &
  (20.6\%$\times m$) &
 (6.7\%$\times m$) &
  (20.6\%$\times m$) &
  (6.7\%$\times m$) &
  (20.6\%$\times m$) &
  (6.7\%$\times m$) &
  (20.6\%$\times m$) &
 (6.7\%$\times m$) &
  (20.6\%$\times m$) &
 (6.7\%$\times m$) &
  (20.6\%$\times m$) \\
KD &
  96.7±5.1 &
  95.3±4.6 &
  76.6±5.9 &
  77.0±8.1 &
  76.0±5.1 &
  78.8±3.2 &
  80.4±3.4 &
  82.0±3.5 &
  82.8±1.6 &
  83.6±1.9 &
  90.4±2.2 &
  91.7±1.9\\
RANDOM &
  67.9±8.0 &
  62.9±8.5 &
  60.1±9.1 &
  61.0±8.5 &
  60.8±9.4 &
  60.2±9.2&
  61.6±5.8 &
  60.2±6.4 &
  57.3±4.3 &
  59.9±3.4 &
  69.6±4.3 &
  64.5±5.6
  \\
GFKD &
  77.8±11.1 &
  72.6±10.4 &
  65.2±7.7 &
  62.1±7.0 &
  61.3±4.0 &
  62.5±3.6&
  67.2±5.5 &
  65.1±5.4 &
  64.7±3.3 &
  64.1±3.0 &
  70.2±3.4 &
  68.1±3.9\\
\textbf{DFAD-GNN} &
  \textbf{87.8±6.9} &
  \textbf{85.6±6.7} &
  \textbf{71.0±3.1} &
  \textbf{69.7±3.5} &
  \textbf{70.0±4.2} &
  \textbf{69.9±5.3} &
   \textbf{73.1±4.3} &
  \textbf{74.9±3.1} &
  \textbf{72.1±2.7} &
  \textbf{71.2±2.0} &
  \textbf{75.4±2.4} &
  \textbf{75.7±2.3}\\ 
{} &
 {(90.8\%$\times t$)} &
  {(88.5\%$\times t$)} &
  {(94.7\%$\times t$)} &
  {(92.9\%$\times t$)} &
  {(89.4\%$\times t$)} &
  {(89.3\%$\times t$)} &
   {(91.3\%$\times t$)} &
  {(93.5\%$\times t$)} &
  {(86.3\%$\times t$)} &
  {(85.3\%$\times t$)} &
  {(81.8\%$\times t$)} &
  {(82.1\%$\times t$)}\\ \hline
Student &
  GCN-5-32 &
  GCN-1-128 &
  GCN-5-32 &
  GCN-1-128 &
  GCN-5-32 &
  GCN-1-128&
  GCN-5-32 &
  GCN-1-128 &
  GCN-5-32 &
  GCN-1-128 &
  GCN-5-32 &
  GCN-1-128\\
  &
  (3.3\%$\times m$) &
  (10.6\%$\times m$) &
 (3.3\%$\times m$) &
  (10.6\%$\times m$) &
 (3.3\%$\times m$) &
  (10.6\%$\times m$) &
 (3.3\%$\times m$) &
  (10.6\%$\times m$) &
 (3.3\%$\times m$) &
  (10.6\%$\times m$) &
 (3.3\%$\times m$) &
  (10.6\%$\times m$) \\
KD &
  86.7±9.4 &
  82.2±10.2 &
  70.9±7.3 &
  70.0±6.9 &
  74.5±3.8 &
  75.5±4.0&
  79.7±3.6 &
  81.1±3.2 &
  81.7±1.1 &
  82.1±2.2 &
  88.2±2.3 &
  87.8±2.4\\
RANDOM &
  58.9±19.3 &
  55.6±21.1 &
  59.4±10.1 &
  55.6±8.1 &
  59.2±8.4 &
  57.9±8.0&
  52.3±2.4 &
  55.1±2.8 &
  55.6±4.4 &
  55.3±5.5 &
  59.3±3.7 &
  57.7±4.2 \\
GFKD &
  70.0±11.2 &
  69.1±10.3 &
  65.0±8.2 &
  61.9±8.5 &
  62.9±7.7 &
  61.4±8.8 &
  63.5±5.3 &
  65.2±5.7 &
  65.7±2.6 &
  64.2±1.9 &
  65.3±2.6 &
  65.1±2.7\\
\textbf{DFAD-GNN} &
  \textbf{74.1±9.3} &
  \textbf{76.4±8.8} &
  \textbf{67.7±2.9} &
  \textbf{67.9±3.5} &
  \textbf{67.2±5.0} &
  \textbf{65.7±3.7} &
  \textbf{69.5±4.8} &
  \textbf{68.6±5.4} &
  \textbf{70.3±1.2} &
  \textbf{69.8±1.8} &
  \textbf{69.9±1.3} &
  \textbf{70.4±1.9}\\
{} &
  {(76.2\%$\times t$)} &
  {(79.0\%$\times t$)} &
  {(90.3\%$\times t$)} &
  {(90.5\%$\times t$)} &
  {(85.8\%$\times t$)} &
  {(83.9\%$\times t$)} &
  {(86.8\%$\times t$)} &
  {(85.6\%$\times t$)} &
  {(84.2\%$\times t$)} &
  {(83.6\%$\times t$)} &
  {(75.8\%$\times t$)} &
  {(76.4\%$\times t$)}\\ \hline
Student &
  GAT-5-32 &
  GAT-1-128 &
  GAT-5-32 &
  GAT-1-128 &
  GAT-5-32 &
  GAT-1-128 &
  GAT-5-32 &
  GAT-1-128 &
  GAT-5-32 &
  GAT-1-128 &
  GAT-5-32 &
  GAT-1-128\\
  &
  (164.6\%$\times m$) &
  (84.5\%$\times m$) &
 (164.6\%$\times m$) &
  (84.5\%$\times m$) &
  (164.6\%$\times m$) &
  (84.5\%$\times m$) &
 (164.6\%$\times m$) &
  (84.5\%$\times m$) &
 (164.6\%$\times m$) &
  (84.5\%$\times m$) &
 (164.6\%$\times m$) &
  (84.5\%$\times m$)   \\
KD &
  87.8±8.9 &
  82.2±10.2 &
  73.2±5.5 &
  69.7±6.8 &
  76.6±3.4 &
  74.5±4.6 &
  80.7±3.0 &
  79.9±2.8 &
  80.3±1.5 &
  81.5±1.2 &
  90.9±1.9 &
  90.6±2.0 \\
RANDOM &
  63.9±17.3 &
  57.5±20.3 &
  60.0±7.1 &
  59.4±6.7 &
  59.8±6.4 &
  60.6±5.6 &
  53.6±4.5 &
  52.9±2.1 &
  56.3±3.6 &
  58.1±3.3 &
  57.6±4.1 &
  55.8±4.3\\
GFKD &
  72.5±13.8 &
  70.4±11.9 &
  63.2±6.5 &
  62.7±7.0 &
  62.2±6.8 &
  62.8±7.9&
  63.7±4.6 &
  64.4±5.2 &
  66.2±2.3 &
  64.9±3.7 &
  67.8±3.5 &
  68.3±4.4 \\
\textbf{DFAD-GNN} &
  \textbf{76.9±6.9} &
  \textbf{77.3±5.9} &
  \textbf{66.4±3.9} &
  \textbf{68.0±4.7} &
  \textbf{67.8±4.9} &
  \textbf{66.0±4.7} &
  \textbf{68.4±3.9} &
  \textbf{68.0±4.7} &
  \textbf{71.1±1.6} &
  \textbf{70.5±2.5} &
  \textbf{73.5±2.6} &
  \textbf{72.3±2.7}\\
{} &
  {(79.5\%$\times t$)} &
  {(79.9\%$\times t$)} &
  {(88.5\%$\times t$)} &
  {(90.7\%$\times t$)} &
  {(86.6\%$\times t$)} &
  {(84.3\%$\times t$)} &
  {(85.4\%$\times t$)} &
  {(84.9\%$\times t$)} &
  {(85.1\%$\times t$)} &
  {(84.4\%$\times t$)} &
  {(79.7\%$\times t$)} &
  {(78.4\%$\times t$)}\\ \hline
Student &
  \begin{tabular}[c]{@{}c@{}}GraphSAGE\\ -5-32\end{tabular} &
  \begin{tabular}[c]{@{}c@{}}GraphSAGE\\ -1-128\end{tabular} &
  \begin{tabular}[c]{@{}c@{}}GraphSAGE\\ -5-32\end{tabular} &
  \begin{tabular}[c]{@{}c@{}}GraphSAGE\\ -1-128\end{tabular} &
  \begin{tabular}[c]{@{}c@{}}GraphSAGE\\ -5-32\end{tabular} &
  \begin{tabular}[c]{@{}c@{}}GraphSAGE\\ -1-128\end{tabular}&
  \begin{tabular}[c]{@{}c@{}}GraphSAGE\\ -5-32\end{tabular} &
  \begin{tabular}[c]{@{}c@{}}GraphSAGE\\ -1-128\end{tabular} &
  \begin{tabular}[c]{@{}c@{}}GraphSAGE\\ -5-32\end{tabular} &
  \begin{tabular}[c]{@{}c@{}}GraphSAGE\\ -1-128\end{tabular} &
  \begin{tabular}[c]{@{}c@{}}GraphSAGE\\ -5-32\end{tabular} &
  \begin{tabular}[c]{@{}c@{}}GraphSAGE\\ -1-128\end{tabular}\\
  &
  (5.9\%$\times m$) &
  (11.1\%$\times m$) &
 (5.9\%$\times m$) &
  (11.1\%$\times m$) &
  (5.9\%$\times m$) &
  (11.1\%$\times m$) &
  (5.9\%$\times m$) &
  (11.1\%$\times m$) &
 (5.9\%$\times m$) &
  (11.1\%$\times m$) &
 (5.9\%$\times m$) &
  (11.1\%$\times m$)  \\
KD &
  87.8±12.1 &
  82.8±9.8 &
  75.6±5.3 &
  70.3±6.6 &
  76.3±3.5 &
  75.7±4.5 &
  80.8±2.6 &
  80.1±2.5 &
  81.2±1.7 &
  81.5±2.5 &
  90.1±1.7 &
  89.5±1.9\\
RANDOM &
  62.2±17.4 &
  57.8±22.7 &
  61.1±7.0 &
  59.9±6.9 &
  57.4±8.5 &
  55.7±6.3 &
  52.6±2.8 &
  53.2±2.9 &
  54.6±3.6 &
  55.5±2.7 &
  54.6±4.5 &
  54.4±4.0 \\
GFKD &
  67.7±12.9 &
  68.1±12.1 &
  62.5±5.9 &
  63.0±6.6 &
  63.3±7.7 &
  61.8±7.9 &
%   GFKD &
  62.3±5.2 &
  63.1±6.0 &
  63.3±2.3 &
  64.7±3.2 &
  63.6±3.8 &
  64.0±3.7\\
\textbf{DFAD-GNN} &
  \textbf{76.5±7.3} &
  \textbf{75.9±6.5} &
  \textbf{66.9±3.7} &
  \textbf{67.5±3.9} &
  \textbf{69.0±6.1} &
  \textbf{67.8±5.4} &
  \textbf{67.5±4.9} &
  \textbf{69.0±3.4} &
  \textbf{68.9±1.1} &
  \textbf{69.6±2.1} &
  \textbf{71.3±3.1} &
  \textbf{69.1±2.9}\\
{} &
  {(79.1\%$\times t$)} &
  {(78.5\%$\times t$)} &
  {(89.2\%$\times t$)} &
  {(90.0\%$\times t$)} &
  {(88.1\%$\times t$)} &
  {(86.6\%$\times t$)} &
  {(84.3\%$\times t$)} &
  {(86.1\%$\times t$)} &
  {(82.5\%$\times t$)} &
  {(83.4\%$\times t$)} &
  {(77.3\%$\times t$)} &
  {(74.9\%$\times t$)}\\ \hline
\end{tabular}}
\caption{Test accuracies (\%) on six datasets. GIN-5-128 means 5 layers GIN with 128 hidden units. (6.7\%$\times m$) under student model means percentage of student model parameters to teacher model parameters, $m$ is the number of teacher model parameters. (90.8\%$\times t$) under DFAD-GNN means percentage of student model accuracy to teacher model accuracy, $t$ is the accuracy of the corresponding teacher network.}
\label{tab:main}
\end{table*}

\subsection{Generator Architecture}

We adopt a generator with fixed architecture for all experiments. The generator takes a 32-dimensional vector sampled from a standard normal distribution $\boldsymbol{z} \sim \mathcal{N}(\mathbf{0}, \boldsymbol{I})$. We process it with a 3-layer MLP of [64,128,256] hidden units respectively, tanh is taken as the activation function. Finally, a linear layer is used to map the 256-dimensional vectors to $N\times T$-dimensional vectors and reshape them as node features $\mathbf{F} \in \mathbb{R}^{N\times T}$. Throughout our experiments, we take the average number of nodes in the training data as $N$ and test the effect of $N$ in the ablation experiment.

\subsection{Teacher/Student Architecture}
To demonstrate the effectiveness of our proposed framework, we consider four GNN models as teacher and student models for a thorough comparison, including: GIN, GCN, GAT and GraphSAGE. Although the GNN model does not always require a deep network to achieve good results, however, from Appendix B, there is no fixed layer and hidden units that can make six datasets achieve the best performance on four different models. %In order to be more unified
For fair comparison, we use 5 layers with 128 hidden units for teacher models. For the student model, we conduct experiments to gradually reduce the number of layers $l\in \left\{ 5,3,2,1 \right\}$ and gradually reduce the number of hidden units $h\in \left\{ 128,64,32,16 \right\}$. We use a graph classifier layer which first builds a graph representation by averaging all node features extracted from the last GNN layer and then passing this graph representation to an MLP.

\subsection{Implementation}
%Our framework is implemented with Pytorch \cite{paszke2017automatic}. 
For training, we use Adam optimizer with weight decay 5e-4 to update student models. The generator is trained with Adam without weight decay. Both student and generator are using a learning rate scheduler that multiplies the learning rate by a factor 0.3 at 10\%, 30\%, and 50\% of the training epochs. 
%The batch size is set to 32 for all datasets. 
The number of updates $k$ of the student model in Algorithm \ref{alg:algorithm} is set to 5. The threshold $\tau$ is empirically set to 0.5.
% \st{To evaluate our framework, we take the prediction accuracy as our metric for graph classification tasks. The details of learning rate for each dataset can be found in the Appendix A.}

\subsection{Baselines}

We compare with the following baselines to demonstrate the effectiveness of our proposed framework. 

\textbf{Teacher:} The given pre-trained model which serves as the teacher in the distillation process.

\textbf{KD:} The generator is removed, and the student model is trained on 100\% original training data in our framework.

\textbf{RANDOM:} The generator's parameters are not updated and the student model is trained on the noisy graphs generated by the randomly initialized generator. 

\textbf{GFKD:} GFKD is a data-free KD for GNNs by modeling the topology of graph with a multinomial distribution~\cite{Deng2021GraphFreeKD}. 

% It first learns the fake graphs that the knowledge in the teacher GNN is more likely to concentrate on, and then uses these fake graphs to transfer knowledge to the student. 

\begin{figure}[t]
    \centering
    \subfigure[PROTEINS]{
        \includegraphics[width=1.6in]{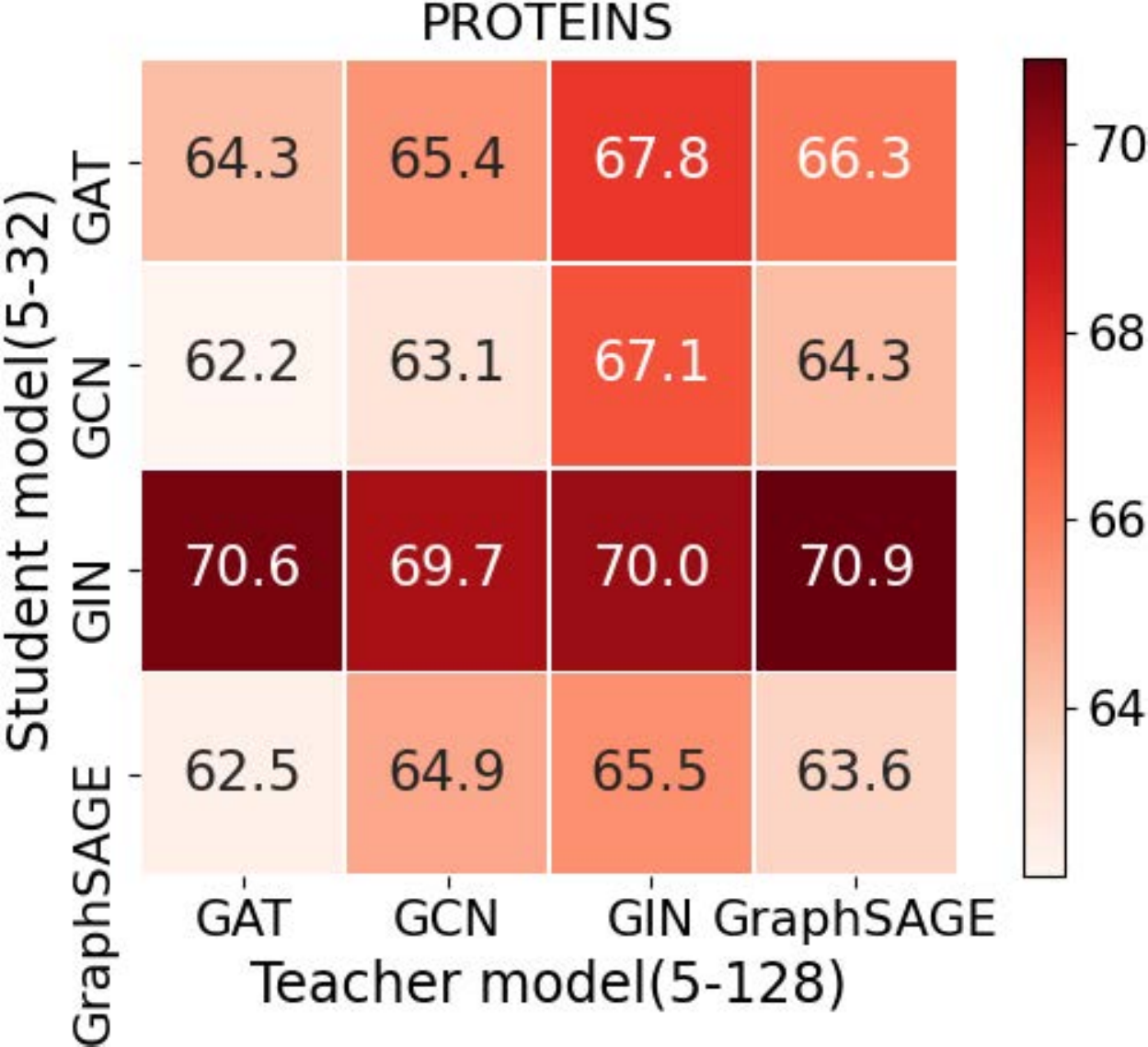}
    }
    \subfigure[COLLAB]{
	\includegraphics[width=1.6in]{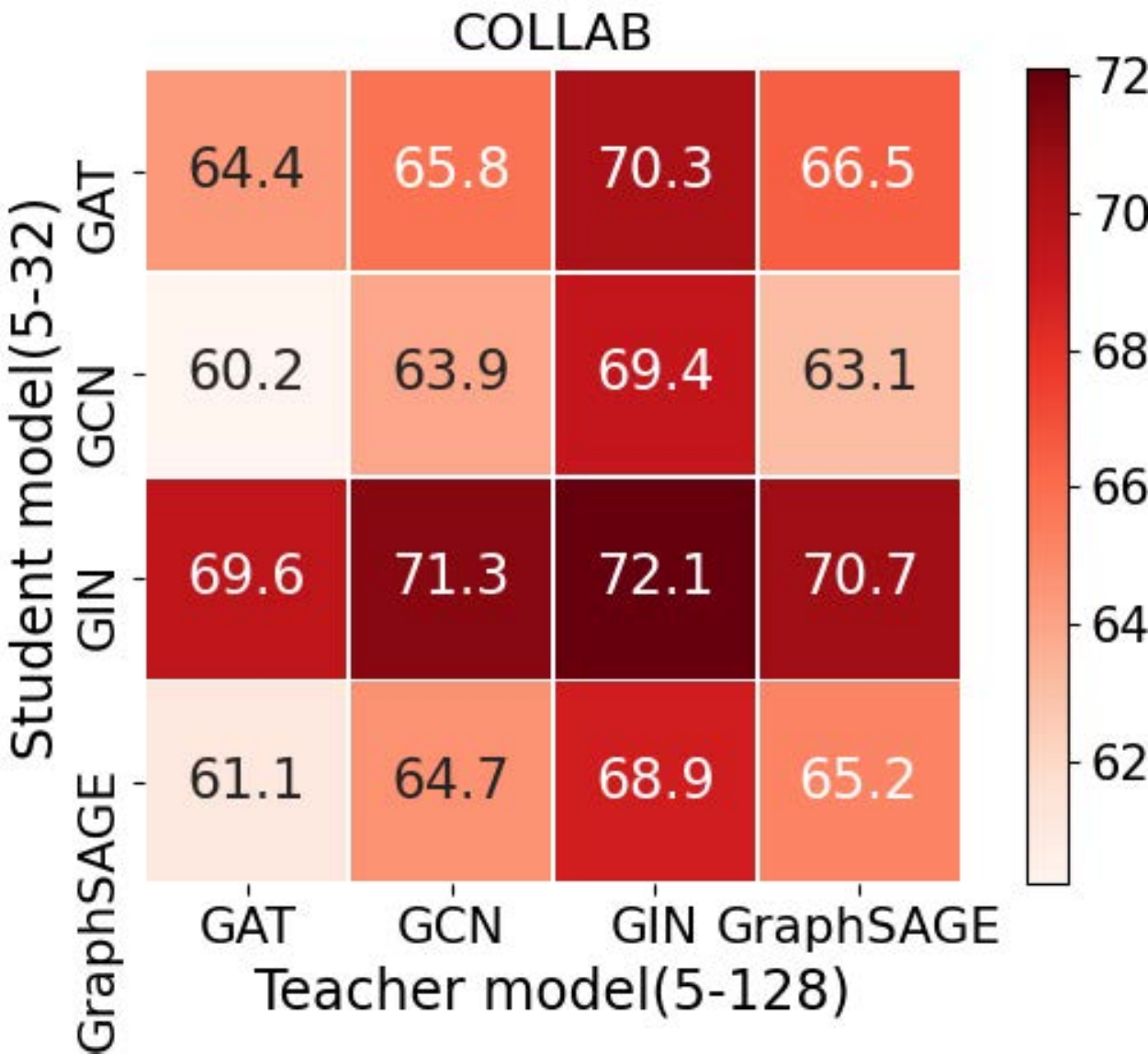}
    }
    \caption{Performance of different teachers to different students on PROTEINS and COLLAB.}
    \label{fig:heat}
\end{figure}

\subsection{Experimental Results}

We have pre-trained all datasets on GCN, GIN, GAT and GraphSAGE with 5 layers and 128 hidden units (5-128 for short), and found that GIN performs best on all datasets (Detailed experimental results can be found in the Appendix B). Therefore, we adopt GIN as the teacher model in Table \ref{tab:main}. We choose two representative architecture 1-128 and 5-32 for four kinds of student models (More experiments with other architectures can be found in the Appendix D). 
% \st{In order to compare the performance of our model across different datasets, we calculate the accuracy percentage of the student model with respect to the teacher model. Besides, we calculate the ratio between the student model's parameters and the teacher model's parameters, i.e., compression ratio.} 

% \begin{figure}[t]
%     \centering
%     \subfigure[Teacher is GIN]{
%         % \includegraphics[width=0.47\columnwidth]{pdf/heat-PROTEINS.pdf}
%         \includegraphics[width=1.57in]{pdf/one-tea.pdf}
%     }
%     \subfigure[Student is GIN]{
% 	\includegraphics[width=1.57in]{pdf/one-stu.pdf}
%     }
%     \caption{Loss convergence on MUTAG (The curve in (a) represents different students, the curve in (b) represents different teachers).}
%     \label{fig:curve}
% \end{figure}

\begin{figure}[t]
\centering
\subfigure[PROTEINS]{
        \includegraphics[width=1.53in]{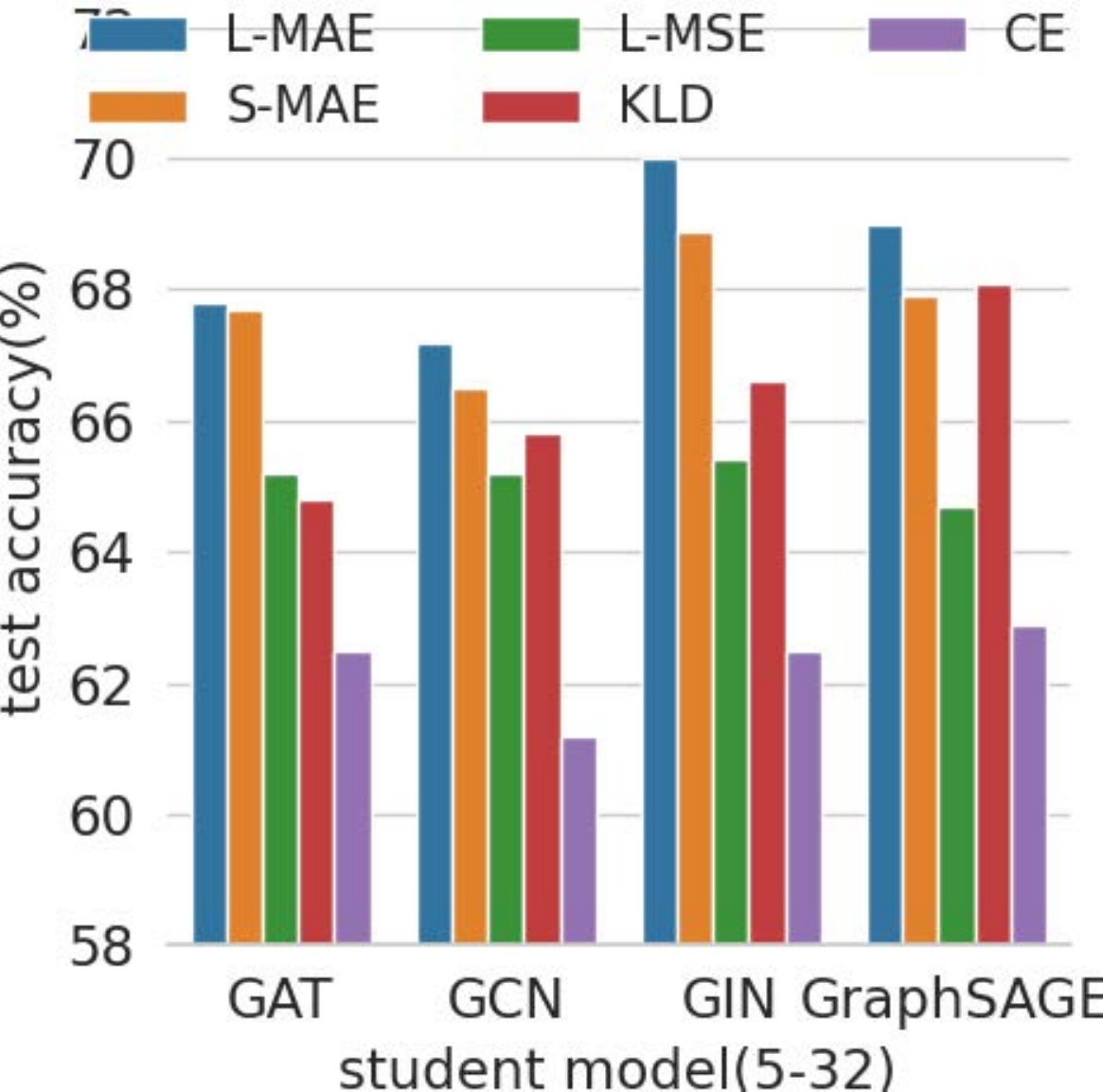}\ \ \
        \vspace{0.01cm}}
\subfigure[COLLAB]{
        \includegraphics[width=1.53in]{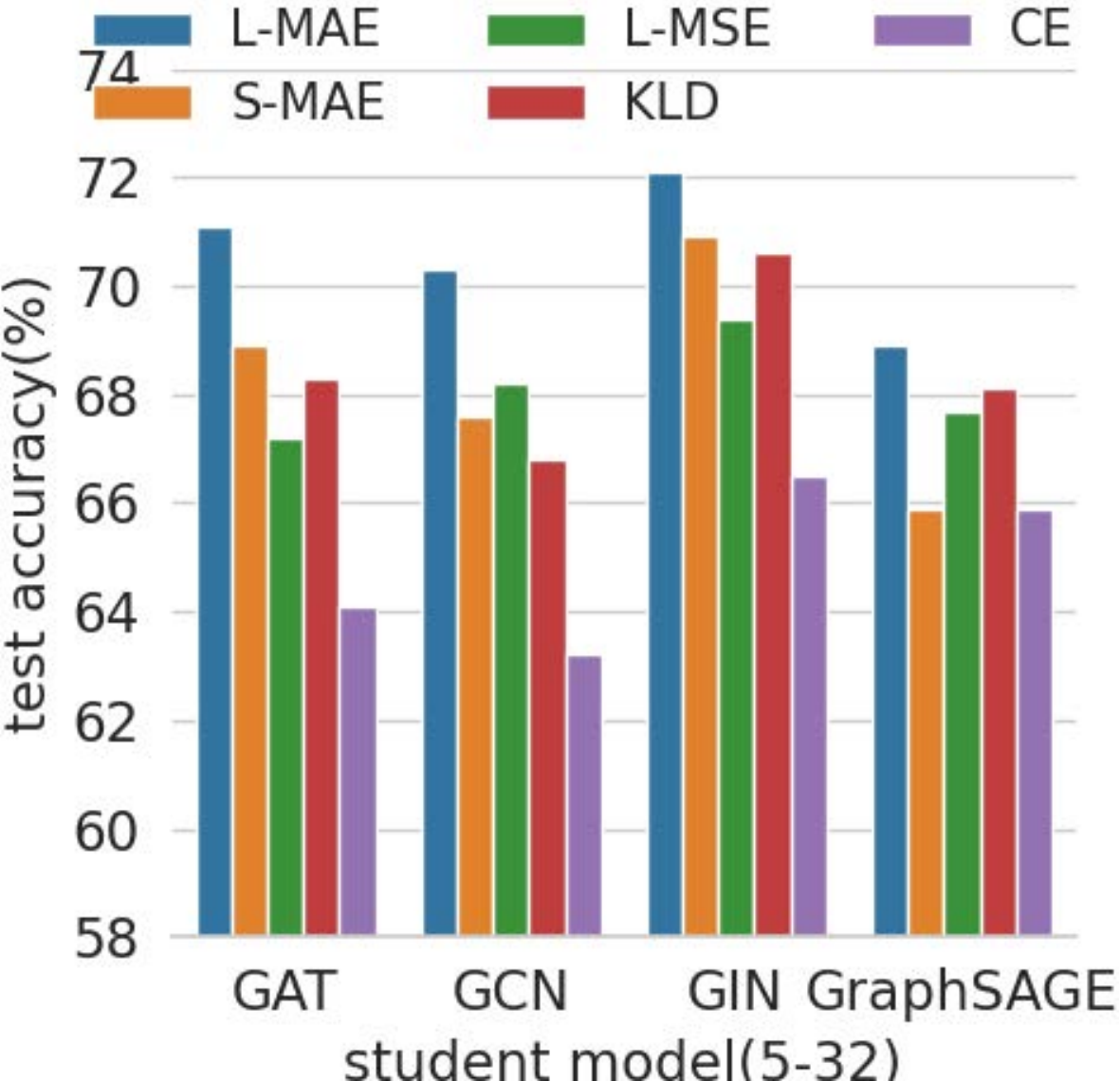}\ \ \
        \vspace{0.01cm}
        %\caption{fig1}
}%
\centering
\caption{Evaluation of different loss functions (teacher model is GIN-5-128).}
\vspace{-10pt}
\label{fig:loss}
\end{figure}

\begin{figure}[t]
    \centering
    \subfigure[PTC\_MR]{
        \includegraphics[width=1.6in]{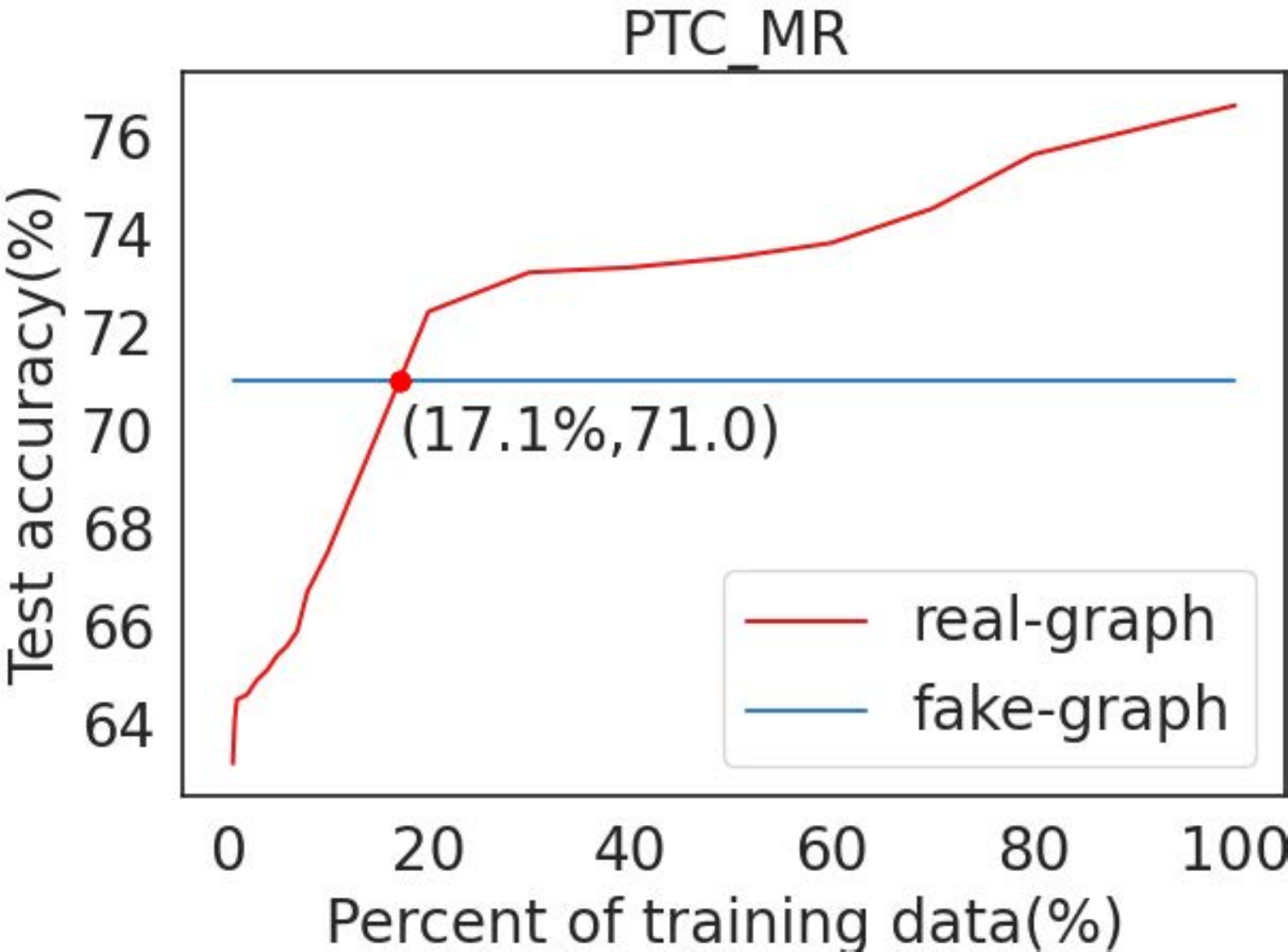}
    }
    \subfigure[IMDB-BINARY]{
	\includegraphics[width=1.6in]{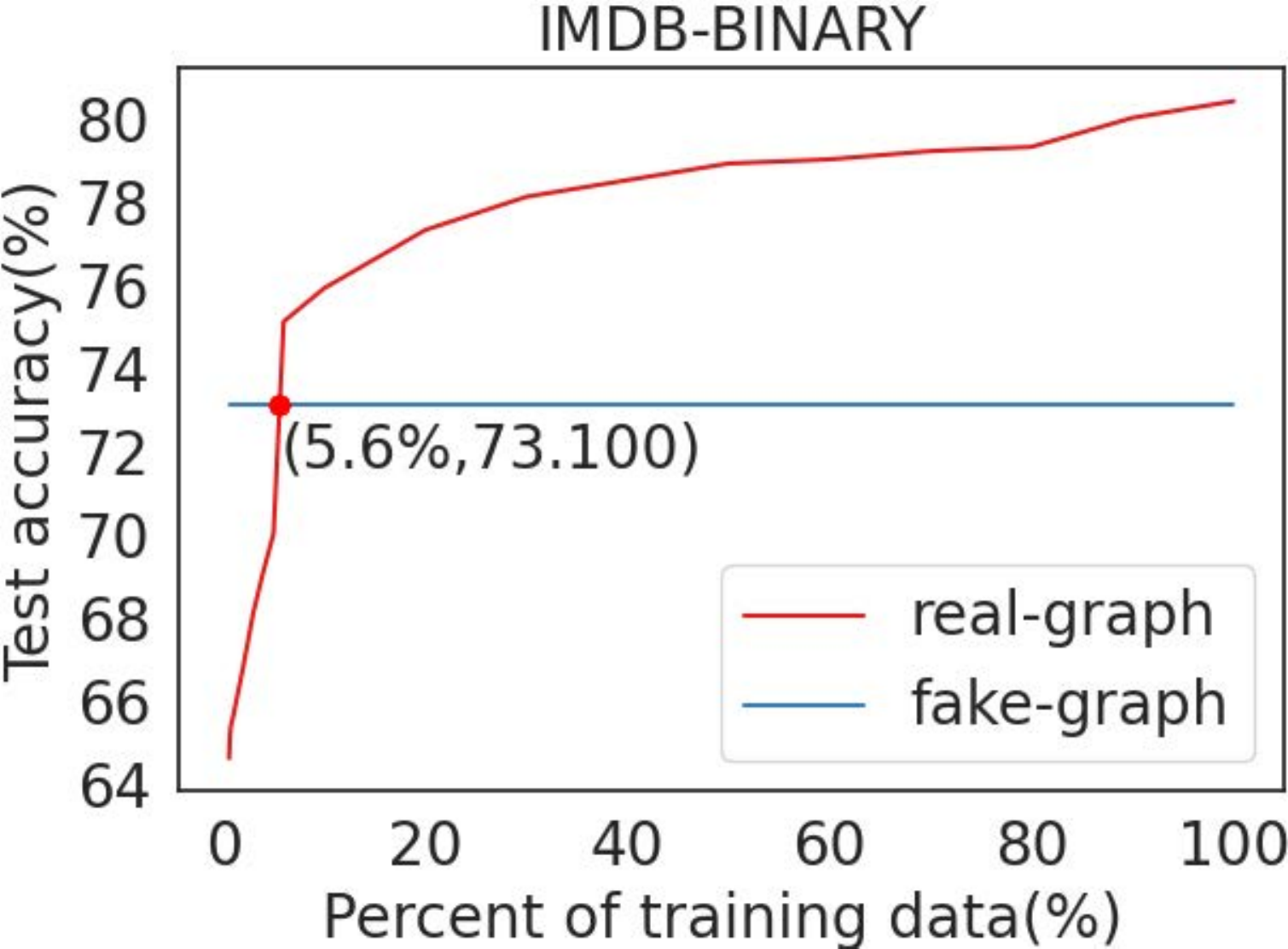}
    }
    \caption{Training with different percentages of real data. The intersection of two lines indicates the needed percentage of the real training data to achieve our data-free performance.}
    \label{fig:trainingdata}
\end{figure}

\begin{figure}[t]
    \centering
    \subfigure[PTC\_MR]{
        \includegraphics[width=1.595in]{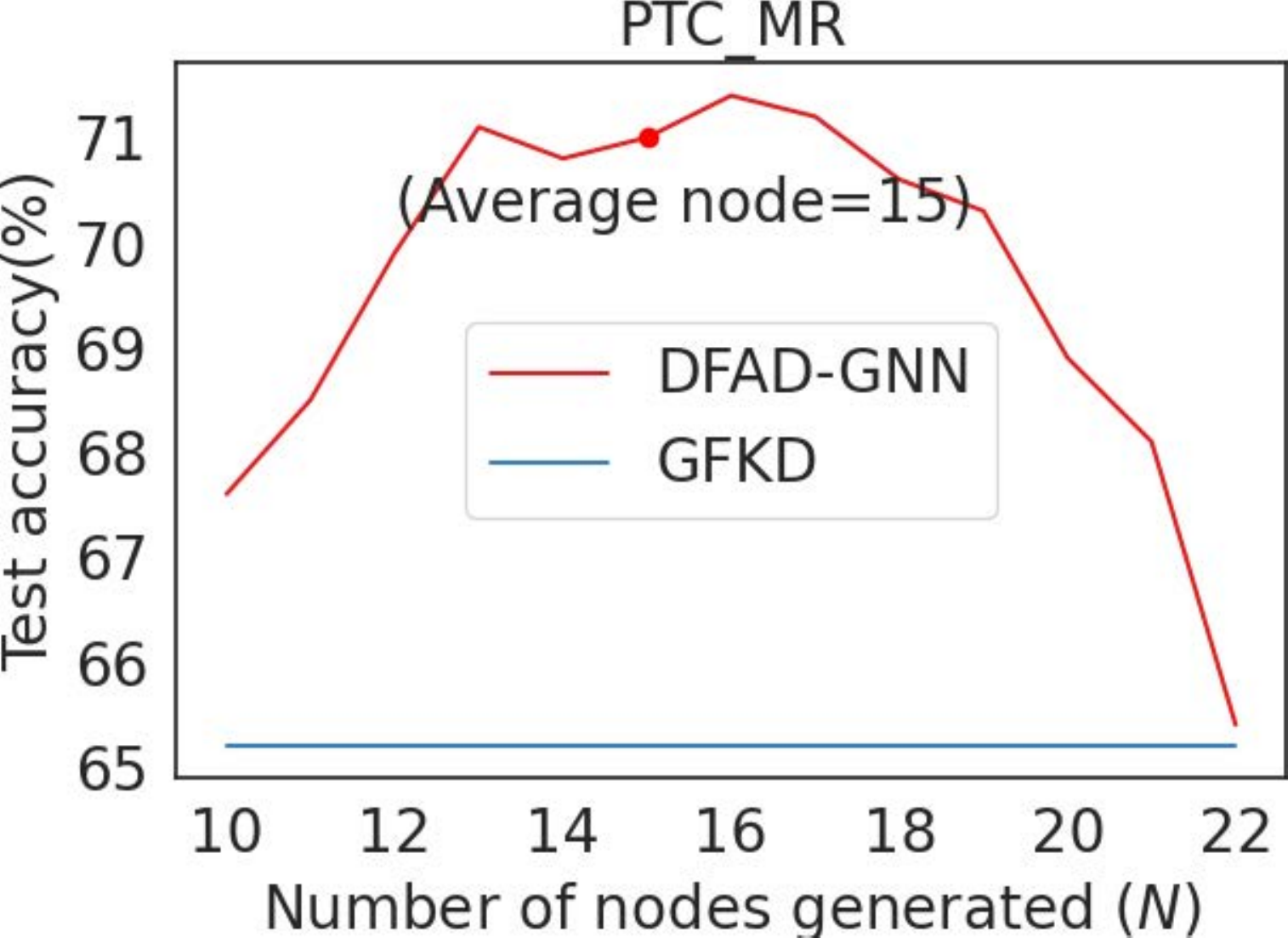}
    }
    \subfigure[IMDB-B]{
	\includegraphics[width=1.6in]{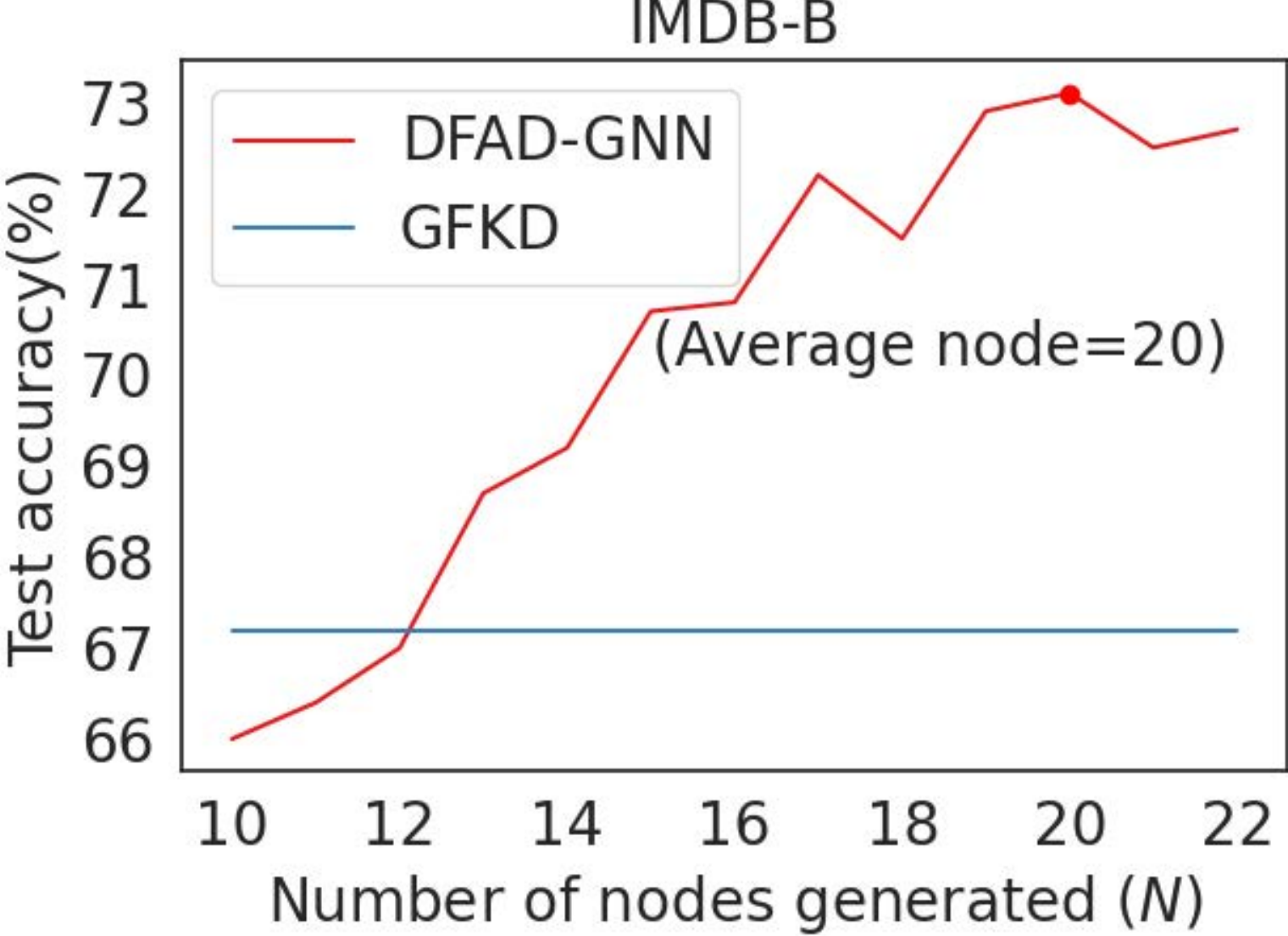}
    }
    \caption{Influence of the node number $N$ (Note that, the GFKD randomly sample the number of nodes from $[10,22]$ for each graph).}
    \label{fig:changeN}
\end{figure}

% \begin{figure*}[t]
%     \centering
%     \subfigure[GAT-5-32]{
%         \includegraphics[width=0.4\columnwidth]{pdf/PTC_GAT-1.pdf}
%     }
%     \subfigure[GCN-5-32]{
% 	\includegraphics[width=0.4\columnwidth]{pdf/PTC_GCN-1.pdf}
%     }
%     \subfigure[GIN-5-32]{
%         \includegraphics[width=0.4\columnwidth]{pdf/PTC_GIN-1.pdf}
%     }
%     \subfigure[GraphSAGE-5-32]{
% 	\includegraphics[width=0.4\columnwidth]{pdf/PTC_GraphSAGE-1.pdf}
%     }
%     \quad    %用 \quad 来换行
%     \subfigure[GAT-5-32]{
%         \includegraphics[width=0.4\columnwidth]{pdf/IMDB_GAT-1.pdf}
%     }
%     \subfigure[GCN-5-32]{
% 	\includegraphics[width=0.4\columnwidth]{pdf/IMDB_GCN-1.pdf}
%     }
%     \subfigure[GIN-5-32]{
%         \includegraphics[width=0.4\columnwidth]{pdf/IMDB_GIN-1.pdf}
%     }
%     \subfigure[GraphSAGE-5-32]{
% 	\includegraphics[width=0.4\columnwidth]{pdf/IMDB_GraphSAGE-1.pdf}
%     }
    
%     \caption{Different percentage of PTC\_MR and IMDB-BINARY on four student models ((a)-(d) is PTC\_MR and (e)-(h) is IMDB-BINARY).}
%     \label{fig:trainingdata}
% \end{figure*}

% \begin{itemize}

From Table \ref{tab:main}, it can be observed that KD's performance is very close to even outperforms the teacher model. That's because KD is a data-driven method which uses the same training data as the teacher model for knowledge distillation. This 
also implies that the loss function of our DFAD-GNN is very effective in distilling knowledge from the teacher model to the student model, as we apply the same loss function in both KD and our DFAD-GNN.
    
We also observe that RANDOM delivers the worst performance as the generator is not updated during the training process, thus the generator will not be able to generate difficult graphs as the student model progresses. Consequently, the student model %did not
fail to learn enough knowledge from the teacher, resulting in poor results.

In terms of the efficacy of our DFAD-GNN, Table \ref{tab:main} shows that our DFAD-GNN consistently outperforms the recent data-free method GFKD \cite{Deng2021GraphFreeKD}. 
% \st{Unlike GFKD that models the graph structures with a multinomial distribution which needs a complicated gradient estimator to optimize, our DFAD-GNN only performs a simple MLP at the generator and uses MAE loss in an end-to-end training process. }
We conjecture the potential reason that DFAD-GNN can significantly outperform GFKD is the teacher encodes the distribution characteristics of the original input graphs under its own feature space. Simply inverting graphs in GFKD tends to overfit to the partial distribution information stored in this teacher model. As a consequence, their generated fake graphs are lacking of generalizability and diversity. In contrast, our generated graphs are more conducive to transferring the knowledge of the teacher model to the student model.
    
In terms of the stability, it can be seen from Table \ref{tab:main} that the standard deviation of our DFAD-GNN is the smallest among all the data-free baselines and across all datasets, indicating that our model can obtain relatively stable prediction results. 

% \st{For the student model, we find that the most expressive architecture is GIN. Because GIN can identify some graph structures that cannot be distinguished by GCN, GAT and GraphSAGE, such as isomorphic graph.}

Another interesting observation is that the performance of the compressed model is not necessarily worse than the more complex model. As can be seen from Table \ref{tab:main}, that performance of a more compressed  student model with 5-32 is not necessarily worse than the student model with 1-128. Therefore, we speculate that the performance of the student model may have no obvious relationship with the degree of model compression, which requires further investigation.

\subsection{Model Analysis}

% \st{\textbf{Loss Convergence}.\\
% %In order to check the model training. 
% Figure 6 shows the loss curves of different student models when teacher model is GIN on MUTAG. Figure 6 shows the loss curves of different teacher models when student model is GIN on MUTAG. It can be seen that as the number of training increases, the loss shows a trend of convergence. Moreover, it is observed that the model converges very quickly, almost within 10 epochs.}

\paragraph{Model Comparison.}
Here, we select PROTEINS and COLLAB with the most training data on molecular datasets and social datasets respectively for cross-training among the four kinds of models. It can be seen from Figure \ref{fig:heat}, when GIN is used as a teacher model, the overall performance of the student model is better, no matter which type of the student model is adopted. 
When GIN is used as a student model, the performance of the student model is significantly improved compared to other student models. We speculate there may be two reasons: (1) Under the same number of layers and hidden units, the GIN model has more parameters than GCN and GraphSAGE, so GIN owns more powerful learning capability. Note that although GAT has more parameters than GIN, it may be better at calculating node attention weights and then performing node classification tasks; (2) GIN is proposed to solve the problem of graph isomorphism. For our small molecule graphs and social network graphs, GIN is more powerful than other models in graph classification tasks. 
% In the Appendix, we provide more experiments for each data set when the teacher is GCN or GIN, and the student is GIN.

\paragraph{Choice of Loss Function.}
% It is essential to keep the balance of the adversarial training. 
% An appropriate adversarial loss should provide stable gradients during training. 
The choice of loss is of paramount importance to a successful distillation. We explore five potential loss functions, including logit-MAE (calculate MAE with pre-softmax activations, L-MAE for short), softmax-MAE (calculate MAE with softmax outputs, S-MAE for short), MSE, KLD, and Cross Entropy (CE for short). These losses are commonly used in the knowledge distillation literature \cite{gou2021knowledge}. Figure \ref{fig:loss} shows that using the MAE achieves significantly better test accuracies compared to other loss functions. Generally speaking, it is better to calculate MAE before softmax, because logits contain more information. As mentioned earlier, when the student model matches more closely to the teacher model, other loss functions tend to suffer from vanishing gradients \cite{fang2019data}. Specifically, backpropagating such vanishing gradients through the generator can harm its learning. To prevent gradients from vanishing, we use the MAE computed with the teacher and student logits. 
% More experimental results on loss comparison can be found in the Appendix D. 

\paragraph{Percentage of Training Data.}
Although in the above reported results, we assume that the student cannot obtain any training data. However, in practice, student may have partial access to the training data. To reflect the practical scenario of knowledge distillation, we conduct extra experiments on PTC\_MR and IMDB-B, by varying the percentage of training data from 0.5\% to 100\%, while keeping other hyper-parameters the same as in the previous experiments. As illustrated in Figure \ref{fig:trainingdata}, PTC\_MR requires 17.1\% of real data to achieve our results, while IMDB-B requires only 5.6\%. This is because PTC\_ MR has less data, thus the percentage of real data required is higher than IMDB-B.
% It can be observed that when the student model is GCN and GIN, more training data is required.

\paragraph{Number of generated nodes.}
To explore the influence of node number $N$ on the model performance, we conduct experiments with varying sizes of $N$ on PTC\_MR and IMDB-B. It can be seen from Figure \ref{fig:changeN} that the model performs better when $N$ takes the value near the average number of nodes in the training set. Far from the average number, the performance will decrease correspondingly due to a large deviation from real data.

%The performance does not always the best when $N$ takes the average value. The main goal of this work is to show the superiority of our method over the baseline methods, rather than finding an optimal $N$.

\begin{figure}
\centering
\subfigure[IMDB-B real graph]{
    \begin{minipage}[t]{0.07\textwidth}
    \includegraphics[width=1.3cm]{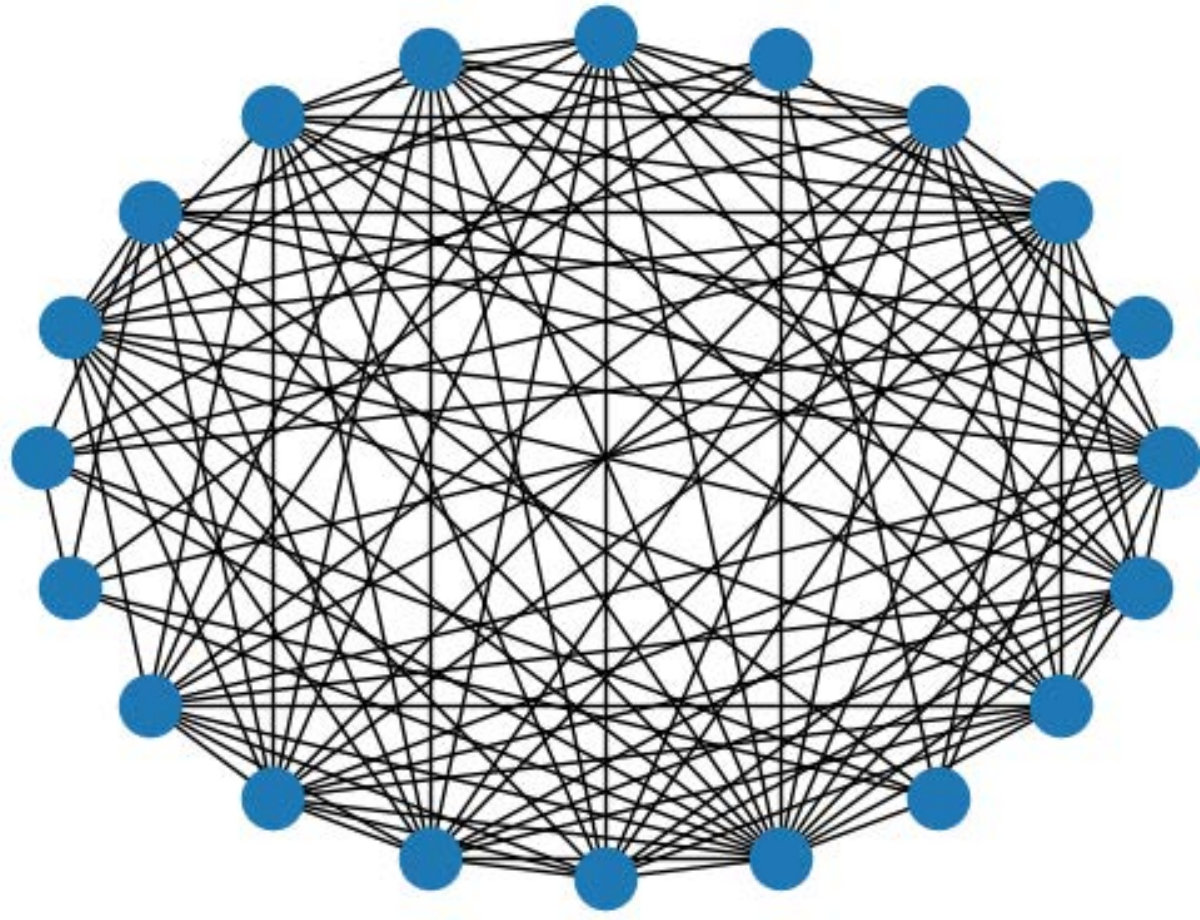}
    \end{minipage}
    \begin{minipage}[t]{0.07\textwidth}
    \includegraphics[width=1.3cm]{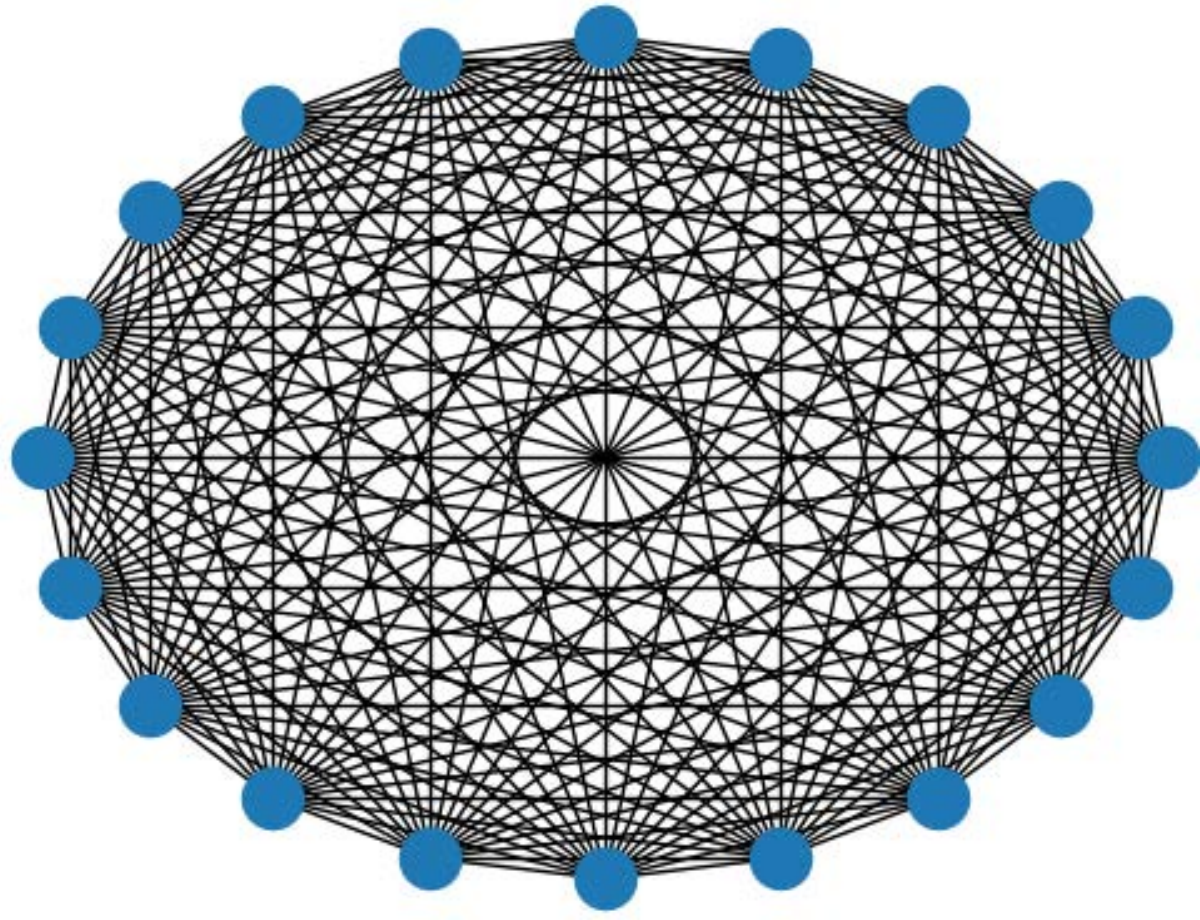}
    \end{minipage}
    \begin{minipage}[t]{0.07\textwidth}
    \includegraphics[width=1.3cm]{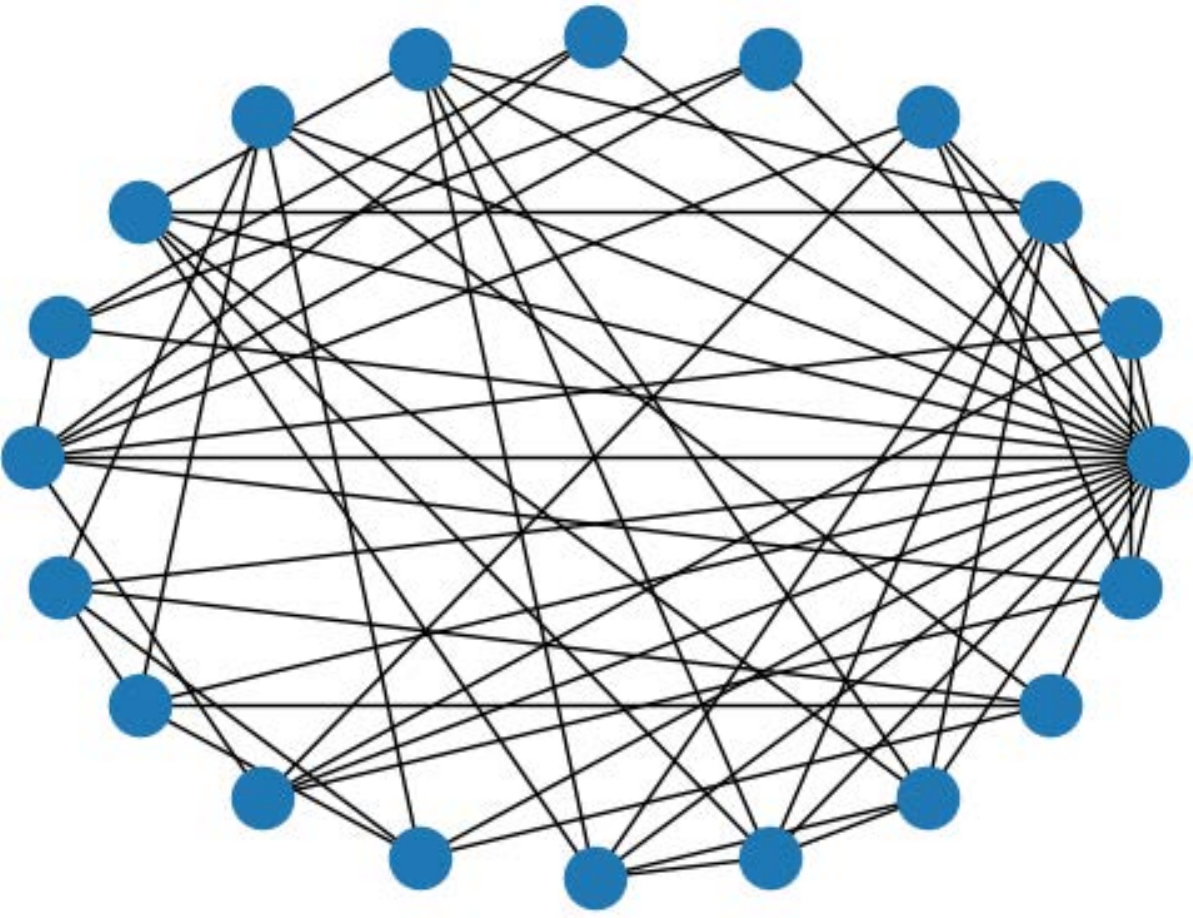}
    \end{minipage}}
\subfigure[PTC\_MR real graph]{
    \begin{minipage}[t]{0.07\textwidth}
    \includegraphics[width=1.3cm]{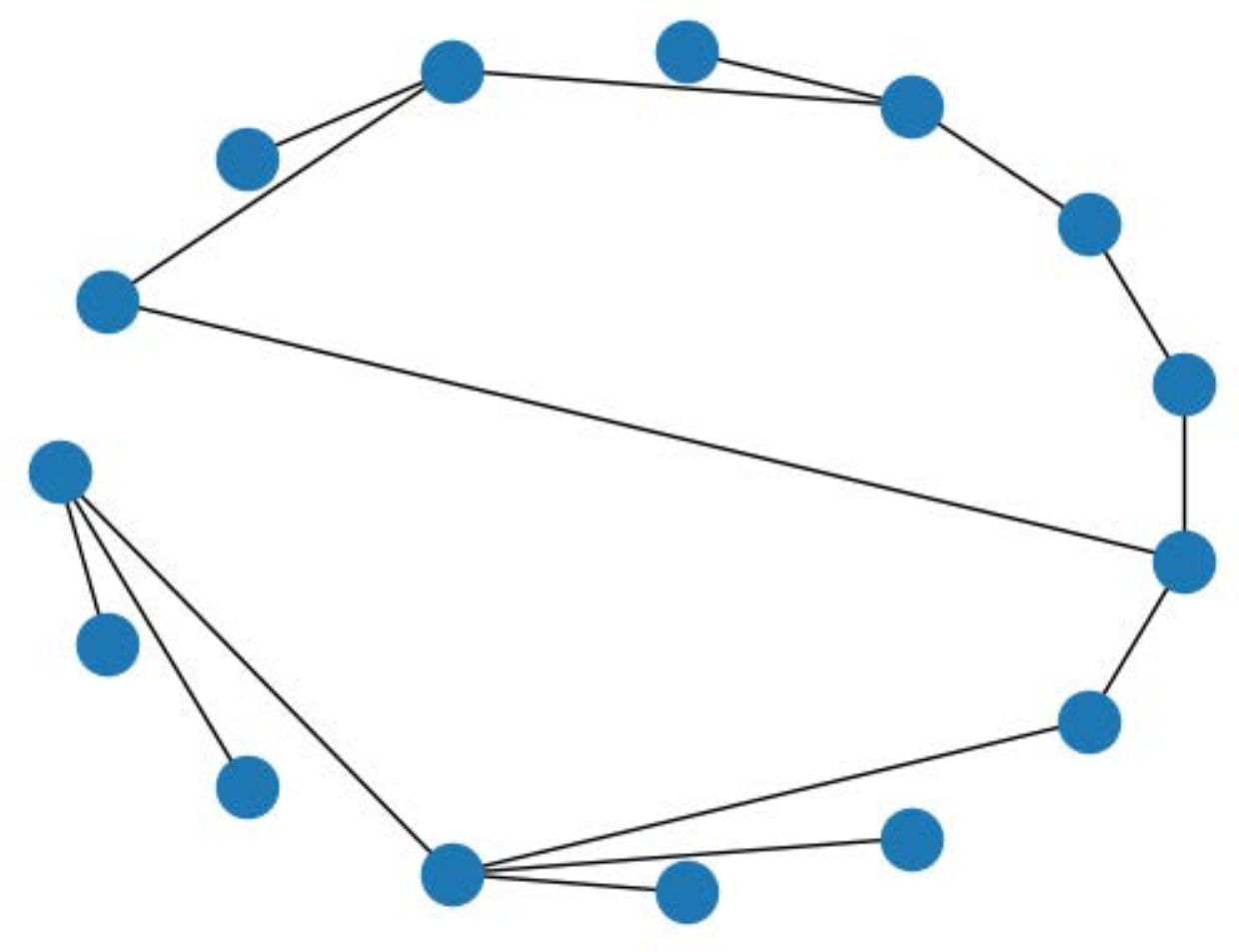}
    \end{minipage}
    \begin{minipage}[t]{0.07\textwidth}
    \includegraphics[width=1.3cm]{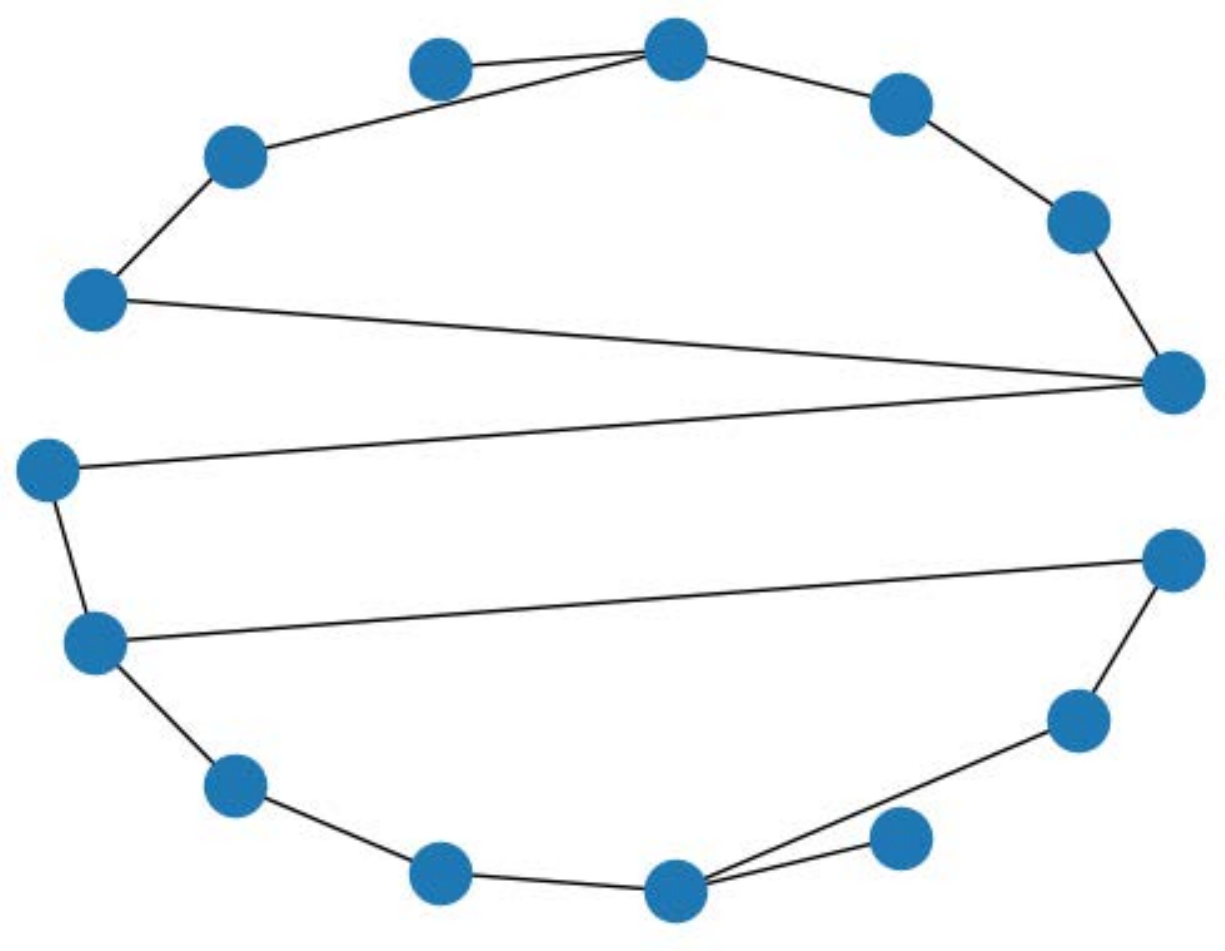}
    \end{minipage}
    \begin{minipage}[t]{0.07\textwidth}
    \includegraphics[width=1.3cm]{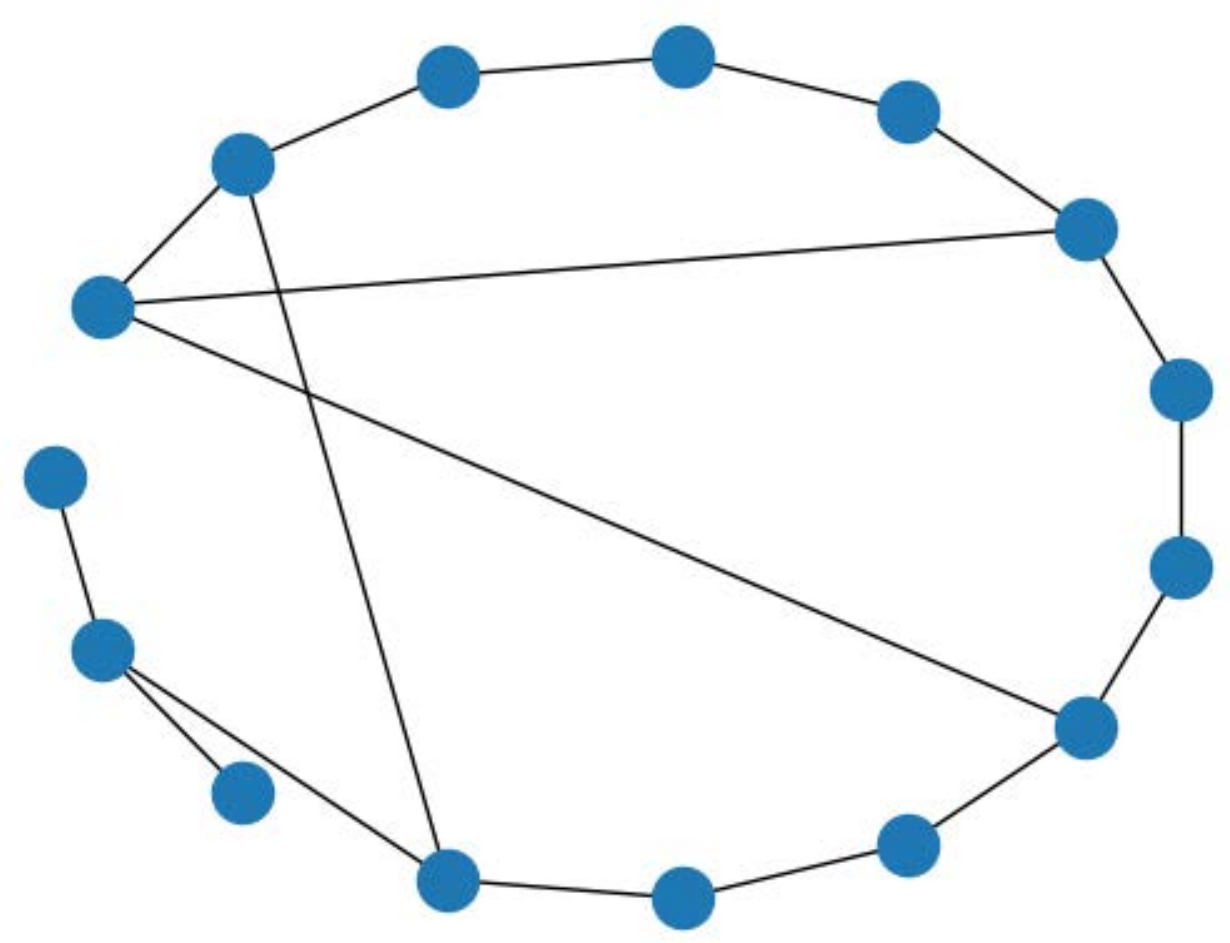}
    \end{minipage}}
    
\subfigure[IMDB-B fake graph]{
    \begin{minipage}[t]{0.07\textwidth}
    \includegraphics[width=1.3cm]{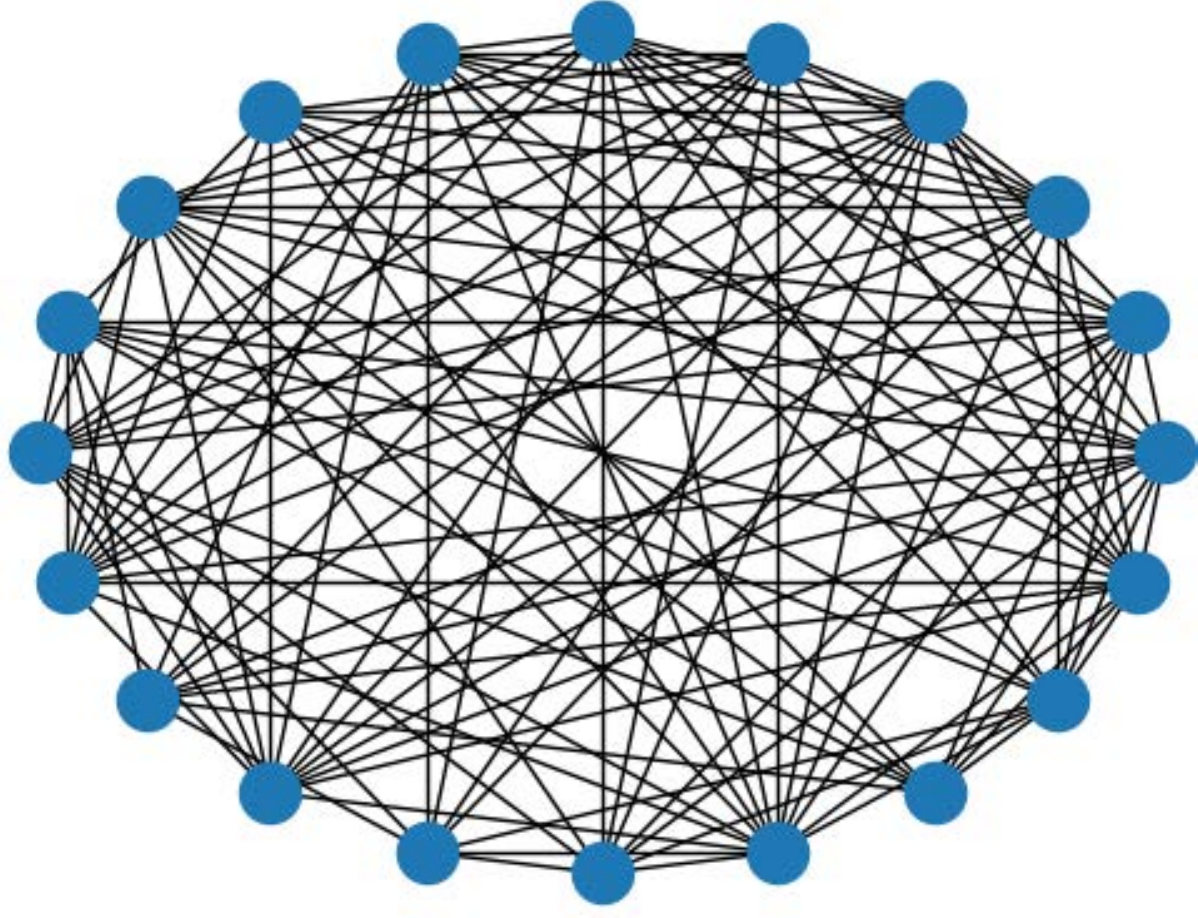}
    \end{minipage}
    \begin{minipage}[t]{0.07\textwidth}
    \includegraphics[width=1.3cm]{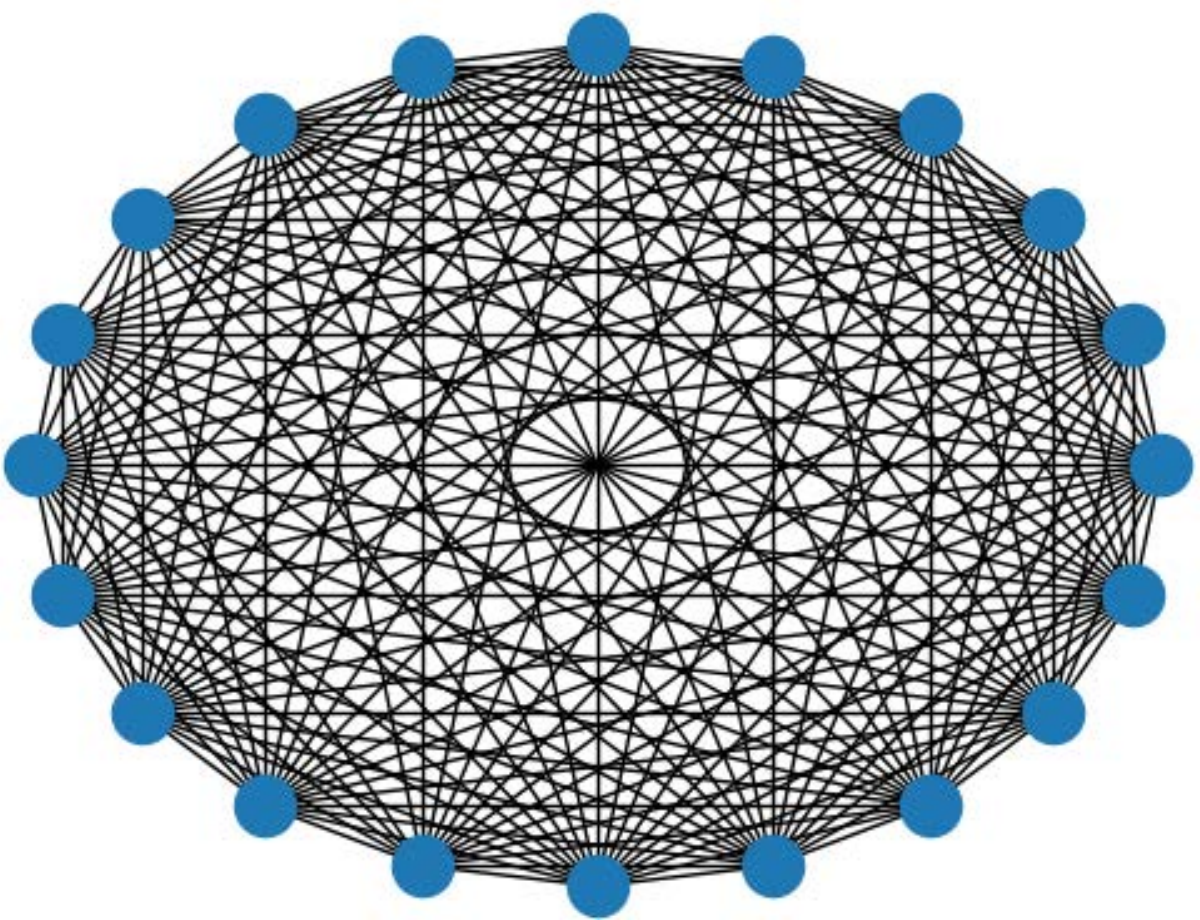}
    \end{minipage}
    \begin{minipage}[t]{0.07\textwidth}
    \includegraphics[width=1.3cm]{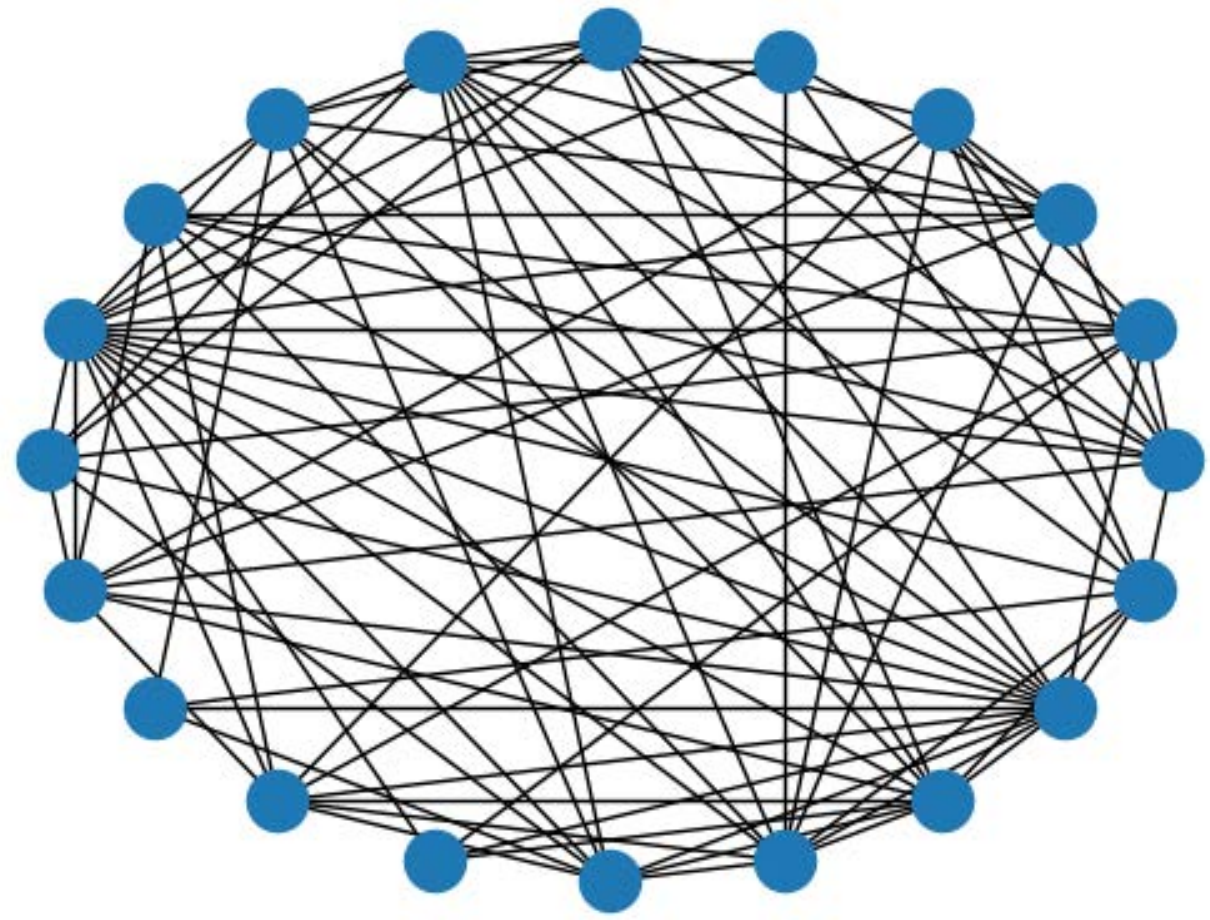}
    \end{minipage}}
\subfigure[PTC\_MR fake graph]{
    \begin{minipage}[t]{0.07\textwidth}
    \includegraphics[width=1.3cm]{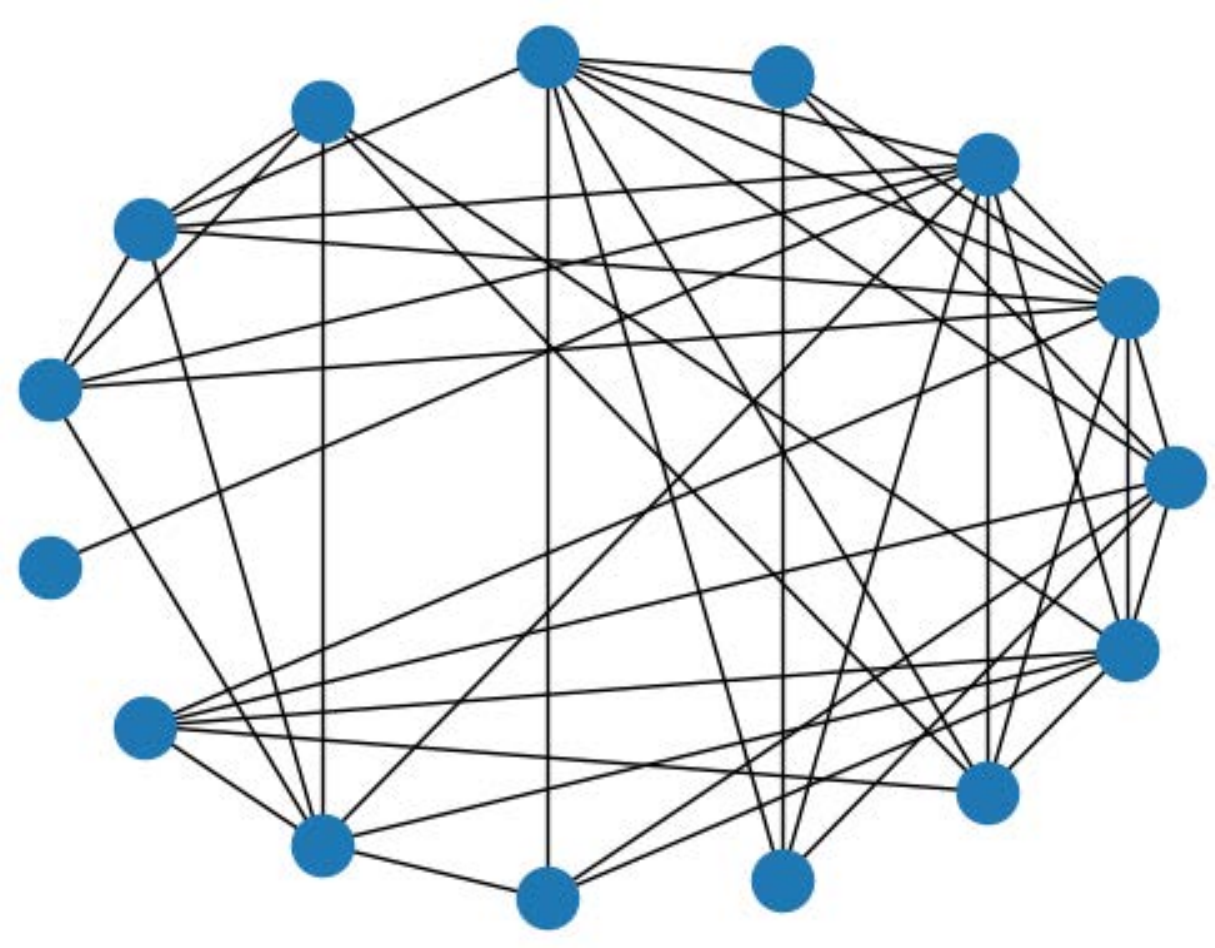}
    \end{minipage}
    \begin{minipage}[t]{0.07\textwidth}
    \includegraphics[width=1.3cm]{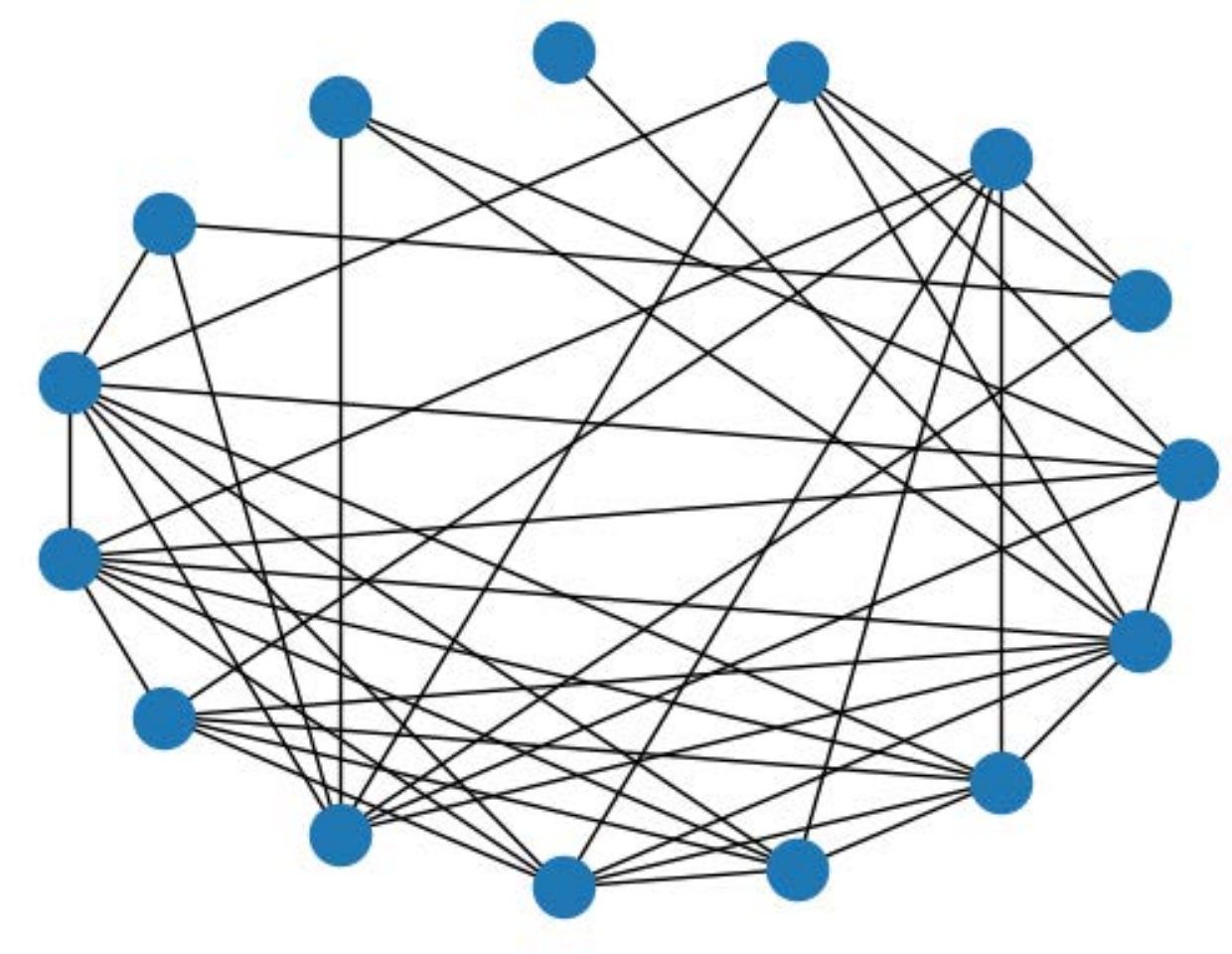}
    \end{minipage}
    \begin{minipage}[t]{0.07\textwidth}
    \includegraphics[width=1.3cm]{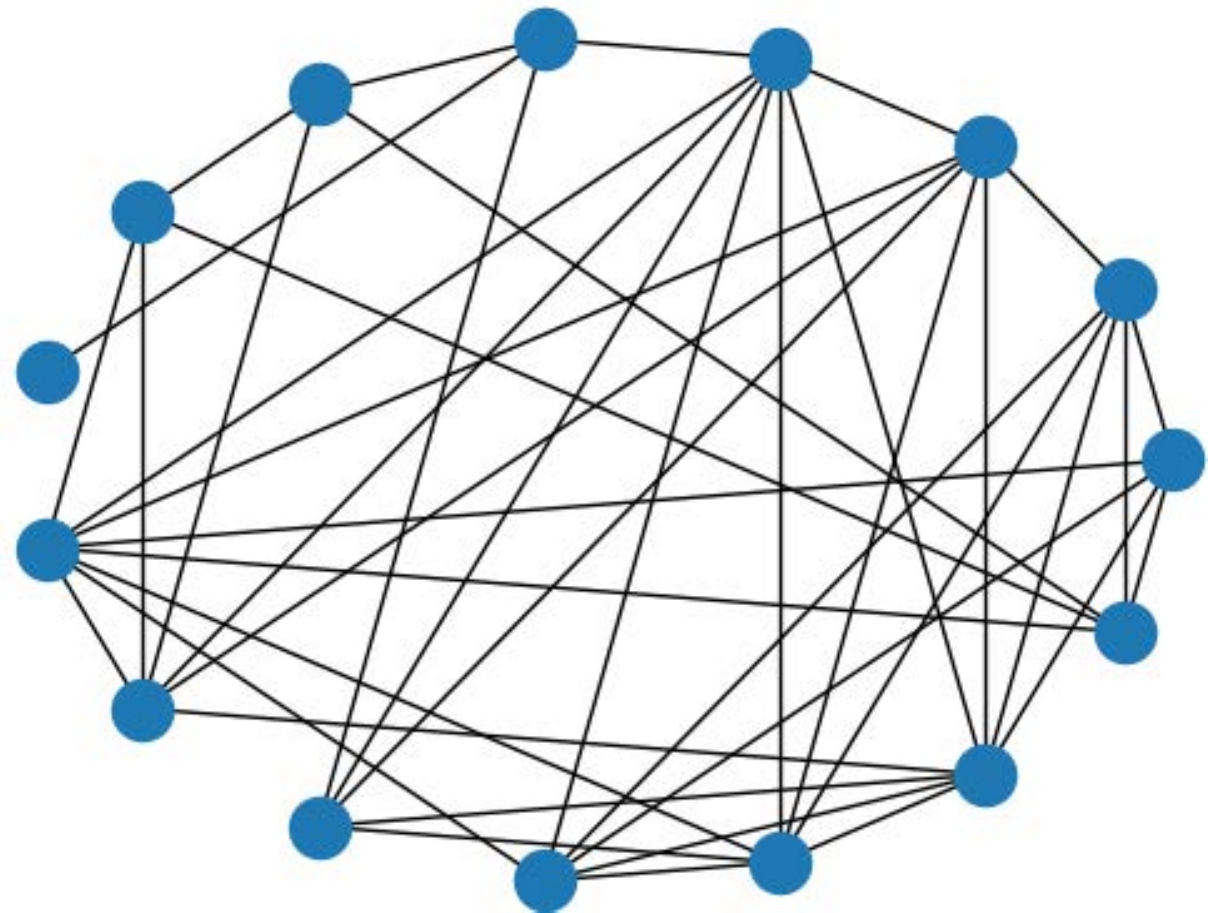}
    \end{minipage}}
\caption{Graph visualization on IMDB-B and PTC\_MR (The first row are real graphs, the second row are fake graphs we generated).}
\label{fig:visual}
\end{figure}

\paragraph{Visualization of Generated Graphs.}
The generated graphs and real graphs of IMDB-B and PTC\_MR are shown in Figure \ref{fig:visual}. Although the generated graphs are not exactly the same as the real graphs, they can be used to generate a student model with relatively good performance.  

\section{Conclusion}

This paper introduces a data-free adversarial knowledge distillation framework on graph neural network for model compression. Without any access to real data, we successfully reduce the discrepancy and obtain a student model with relatively good performance. Our extensive experiments on graph classification demonstrate that our framework can be effectively applied to different network architectures. 
% To the best of our knowledge, it is also the first end-to-end data-free framework for graph knowledge distillation. 
In the future, we will extend this work to multi-teacher scenarios and continue to explore how to generate more complex graphs under various generator structures.

\section*{Acknowledgments}

This work is supported in part by the National Natural Science
Foundation of China (No. 62192784, U20B2045, 62172052, 61772082, 62002029, U1936104). It is also supported in part by The Fundamental Research Funds for the Central Universities 2021RC28.
% \clearpage

\bibliographystyle{named}
\bibliography{ijcai22}
\end{document}